\documentclass[pdflatex,sn-mathphys-num, hidelinks, iicol]{sn-jnl}

\usepackage[switch]{lineno}
\usepackage{amsmath,amsfonts}
\usepackage{algorithmic}
\usepackage{algorithm}
\usepackage{array}
\usepackage[caption=false,font=normalsize,labelfont=sf,textfont=sf]{subfig}
\usepackage{textcomp}
\usepackage{stfloats}   
\usepackage{url}
\usepackage{verbatim}
\usepackage{graphicx}
\usepackage[table]{xcolor}
\usepackage{threeparttable}
\usepackage{graphicx}
\usepackage{amsmath}
\usepackage{amssymb}
\usepackage{booktabs}
\usepackage{multirow}
\usepackage{diagbox}
\DeclareMathOperator*{\argmax}{arg\,max}

\definecolor{lightgray}{gray}{0.92}
\definecolor{newgray}{rgb}{.6, .6, .6}

\usepackage[numbers]{natbib}
\usepackage{hyperref}       
\usepackage[capitalize]{cleveref}

\theoremstyle{thmstyleone}%
\theoremstyle{thmstyletwo}%

\theoremstyle{thmstylethree}%

\begin{document}

\title[Article Title]{PartImageNet++ Dataset: Enhancing Visual Models with High-Quality Part Annotations}


\author[1]{\fnm{Xiao} \sur{Li}}\email{xiaoli.cst@gmail.com}

\author[2]{\fnm{Zilong} \sur{Liu}}\email{zilongliu@mail.sdu.edu.cn}

\author[3]{\fnm{Yining} \sur{Liu}}\email{lyn851037220@163.com}

\author[4]{\fnm{Zhuhong} \sur{Li}}\email{zl425@duke.edu}

\author[5]{\fnm{Na} \sur{Dong}}\email{3220220911@bit.edu.cn}

\author[3]{\fnm{Sitian} \sur{Qin}}\email{qinsitian@hitwh.edu.cn}

\author*[1,6]{\fnm{Xiaolin} \sur{Hu}}\email{xlhu@mail.tsinghua.edu.cn}

\affil[1]{Department of Computer Science and Technology, BNRist, Institute for Artificial Intelligence, IDG/McGovern Institute for Brain Research, Tsinghua University, Beijing, China}

\affil[2]{School of Computer Science and Technology, Shandong University, Qingdao, China}

\affil[3]{Harbin Institute of Technology, Weihai, China}
\affil[4]{Duke University, Durham, NC, USA}
\affil[5]{Beijing Institute of Technology, Beijing, China}
\affil[6]{Chinese Institute for Brain Research, Beijing, China.}


\abstract{To address the scarcity of high-quality part annotations in existing datasets, we introduce PartImageNet++ (PIN++), a dataset that provides detailed part annotations for all categories in ImageNet-1K. With 100 annotated images per category, totaling 100K images, PIN++ represents the most comprehensive dataset covering a diverse range of object categories. Leveraging PIN++, we propose a Multi-scale Part-supervised recognition Model (MPM) for robust classification on ImageNet-1K. We first trained a part segmentation network using PIN++ and used it to generate pseudo part labels for the remaining unannotated images. MPM then integrated a conventional recognition architecture with auxiliary bypass layers, jointly supervised by both pseudo part labels and the original part annotations. Furthermore, we conducted extensive experiments on PIN++, including part segmentation, object segmentation, and few-shot learning, exploring various ways to leverage part annotations in downstream tasks. Experimental results demonstrated that our approach not only enhanced part-based models for robust object recognition but also established strong baselines for multiple downstream tasks, highlighting the potential of part annotations in improving model performance. The dataset and the code are available at \url{https://github.com/LixiaoTHU/PartImageNetPP}.}




\maketitle
\section{Introduction}\label{sec1}
Several renowned cognitive psychological theories \cite{RBC, xinlixue2}, such as the recognition-by-components theory \cite{RBC}, propose that humans decompose objects into parts, considering hierarchical representations and spatial relationships for recognition. Empirical evidence \cite{infants, opc, attentioneye, oldrbc} from psychology supports this theoretical framework. 
Drawing on these theoretical insights, recent research has been conducted to explore how the information in parts of objects can be utilized to assist deep neural networks in performing various computer vision tasks. For example, segmenting objects into their constituent parts plays an important role in both object segmentation and part segmentation tasks \cite{partseg,objparsing,PartWhole,mambapw,token}. Furthermore, empirical studies have demonstrated that effectively utilizing part information in few-shot learning tasks can enhance model performance \cite{fewshot1, fewshot2, fewshot3, DMPCL}. In addition, a part-based recognition model has been proposed to enhance adversarial robustness in recognition \cite{pinpp}. 
However, its effectiveness is limited to small-scale, non-standard datasets due to the scarcity of part annotations. 

\begin{figure}[!t]
  \centering
   \includegraphics[width=\linewidth]{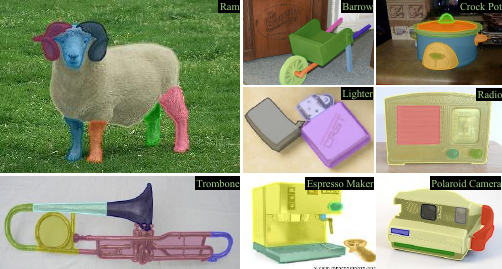}
   \caption{Examples of annotated images in PIN++. The names of the objects are displayed in the upper-right corner of each image, while the part names are not revealed in this context.}
   \label{fig:dataset}
\end{figure}

To address this issue, we introduce PartImageNet++ (PIN++ for short), a dataset that provides high-quality part segmentation annotations for all categories within the widely utilized ImageNet-1K \cite{imagenet} (referred to as IN-1K). Our detailed annotation scheme ensures the quality of part annotations, with 100 images per category annotated, which totaled 100K images with part annotations. All annotations were manually created to avoid bias from auxiliary models. PIN++ is, to our best knowledge, the most extensive dataset offering high-quality part segmentation annotations across the most diverse range of object categories, including creatures, artifacts, rigid, and non-rigid objects. \cref{fig:dataset} showcases some examples of annotated images. We introduce this dataset in detail in \cref{sec:dataset} and show that existing techniques fail to obtain high-quality part segmentation results without such annotations.


Before PIN++, the most extensive dataset, PartImageNet \cite{partimagenet} (PIN for short), encompasses approximately 24K images in 158 categories selected from the large-scale IN-1K \cite{imagenet}. PIN exhibits several limitations, mainly its reduced size compared to the standard IN-1K dataset. Its concentration on animal classes potentially limits its representativeness for the diversity required in general recognition tasks. These constraints impede direct comparative analysis of part-based models \cite{rock, carlinipart} trained on PIN with state-of-the-art (SOTA) methods trained on IN-1K.

With the help of detailed part annotations in the PIN++ dataset, we also developed part annotation-based methods for robust classification on IN-1K. Initially, we generated pseudo part labels for other unannotated IN-1K images, by training a part segmentation network using the PIN++ annotations. Subsequently, a new Multi-scale Part-supervised recognition Model (MPM) was designed to better leverage part annotations without any increase in parameters or computational load during the inference process. 
Under the supervision of both the pseudo part labels and the detailed part annotations in PIN++, we integrated the classical recognition architecture with auxiliary bypass layers. This integration produced a more robust intermediate representation, enabling MPM to leverage part annotations better. The experimental results demonstrate that, when combined with adversarial training (AT) \cite{at}, it achieved better adversarial robustness on the large-scale IN-1K, outperforming strong AT baselines in both the seen and unseen adversarial threats. Furthermore, MPM exhibited improved robustness on common image corruptions \cite{imagenetc} and several out-of-distribution (OOD) datasets. In addition, MPM achieved higher alignments with human vision.

In addition to designing the MPM, we also established benchmarks for three downstream tasks (part segmentation, object segmentation, and few-shot learning) on the PIN++ dataset. To utilize the part information, we designed specific methods for the three downstream tasks and achieved better results than those without the part information. 

 

Part and object segmentation tasks serve two different goals: to segment specific components of objects and to delineate entire object instances in their entirety. For part segmentation, we implemented multiple variants of Mask R-CNN \cite{maskrcnn}—each configured with different backbone networks—on the PartImageNet++ dataset, and we derived a series of noteworthy findings from these experiments. For object segmentation, we carried out a comparative analysis to evaluate how model performance differs with the inclusion versus exclusion of part-level annotations. Our experimental results demonstrate that integrating part-level details enables more precise delineation of object boundaries. This improvement stems from the unique advantage of part-based models: they can capture fine-grained local features and structural patterns that are typically overlooked when an object is treated as a single, holistic unit.

The goal of few-shot learning is to learn a classification rule from a very small number of labeled examples. Within this paradigm, we observed that part-based representations can serve as a potent inductive bias, enabling models to achieve stronger generalization on novel classes when training data is limited. Architecturally, integrating few-shot learning frameworks with dedicated part branches facilitates targeted localized feature extraction, which in turn boosts both the model’s accuracy and interpretability. Consequently, the resulting model not only delivers superior performance but also offers transparent insights into how each component contributes to its final prediction outcomes.

In summary, our contributions are as follows:
\begin{itemize}
    \item We created PIN++, an extensive dataset enriched with part annotations. It is poised to advance research in part-based models for robust object recognition and other visual tasks centered around part understanding.
    
    \item We introduced MPM, a new part-supervised model for robust recognition, designed to optimize the utilization of part annotations. This approach delivers enhanced robustness across a range of scenarios and evaluation metrics, without incurring additional computational costs during inference.
    
    \item We established a comprehensive suite of benchmarks for downstream tasks, including segmentation and few-shot learning based on PIN++. We further explored how to better align model behavior with human recognition mechanisms to boost performance via part annotations.
\end{itemize}

This manuscript is an extension of our preliminary conference paper~\cite{pinpp}. In the preliminary version, we primarily focused on the construction of the initial dataset and validated the efficacy of part-based models in robust image classification. Building upon this foundation, this paper presents substantial extensions in terms of the richness of part-based visual tasks, tailored model designs for different scenarios, and a comprehensive analysis of dataset annotations and experimental insights:
\begin{itemize}
    \item We explore the transferability of part information to data-scarce scenarios. We investigate methodologies for rapidly adapting existing models to part-based datasets. Specifically, we propose incorporating auxiliary part branches into standard few-shot architectures, which improves classification accuracy by leveraging fine-grained part cues.
    
    \item Moving beyond image-level classification, we establish a series of benchmark experiments on PIN++ for dense prediction, including part segmentation and object segmentation. The experimental results demonstrate that incorporating part annotations consistently enhances model performance across these complex visual tasks.

    \item We conduct a rigorous comparative analysis of the dataset and previous experimental results. Through quantitative evaluation and visualization of part annotations (comparing PIN and PIN++), we verify the superior annotation quality and consistency of PIN++. 
\end{itemize}

The rest of the paper is organized as follows:  \cref{sec:RelatedWork} reviews related work. \cref{sec:dataset} details the construction of PIN++ and provides relevant discussions. \cref{sec:MPM} introduces the MPM model and evaluates its robustness. \cref{sec:ExperimentandBenchmark} and \cref{sec:fewshot} present the benchmark evaluations for object segmentation tasks and few-shot learning tasks. Finally, \cref{sec:conclusion} concludes the paper.

\section{Related Work}
\label{sec:RelatedWork}
\subsection{Part Datasets}

The concept of \textit{parts} holds significant importance in both human visual cognition \cite{RBC} and computer vision domains. To conduct research on part-related visual tasks, the availability of part datasets is crucial. However, annotating object parts is both challenging and expensive, and most part datasets are only limited to specific domains, such as \textit{cars} \cite{wang2015unsupervised, CarFusion, Apollocar3D} and \textit{humans} \cite{ATR, LIP, MHP, CIHP}. Furthermore, existing part datasets targeting common objects \cite{Cityscapes-Panoptic-Parts, PASCAL-Part, ade20k, pinpp, paco} often have a limited number of object categories. Cityscapes PanopticParts (Cityscapes PP.) \cite{Cityscapes-Panoptic-Parts}, Pascal-Part \cite{PASCAL-Part}, and ADE20K \cite{ade20k} provide part annotations for 5, 20, and 80 object categories, respectively. Although the recent PACO dataset \cite{paco} offers a large number of part annotations, it covers only 75 object categories. PIN \cite{partimagenet}, containing 158 categories from IN-1K \cite{imagenet}, represents the dataset with the most object categories in terms of part annotations. 



\subsection{Part and Object Segmentation}
Modeling objects in terms of parts has a long history in computer vision \cite{parthis, his1, his2}. The part segmentation task aims to segment specific parts of an object from an image, such as accurately separating parts such as the wheels of a car, the wings of an aircraft, etc. This task is crucial for understanding the structure and function of objects and is widely used in areas such as robot operation~\cite{UDS, Mobile-Seed}. The goal of the object segmentation task is to segment complete objects from an image, which focuses more on the overall boundaries and shapes of the objects than partial segmentation. Object segmentation has a wide range of applications in computer vision, such as autonomous driving~\cite{3D-SeqMOS, autoobj1}. 



\subsection{Adversarial Robustness}
Szegedy et al. \cite{adv13} first revealed the vulnerability of Deep Neural Networks (DNNs) to adversarial examples. Since then, numerous methods have been proposed to enhance the adversarial robustness of DNNs, whereas most of them have been rendered ineffective by new adaptive attacks \cite{AthalyeC018, adaptive20}. Among them, AT has emerged as the de facto paradigm for training adversarially robust DNNs \cite{liadbm, li2025pbcat}. Early AT works \cite{trades, awp}  primarily focused on small-scale, low-resolution datasets such as CIFAR-10 \cite{cifar10}. Recently, there has been a growing interest in investigating AT on IN-1K \cite{imagenet}, as this dataset serves as a standard benchmark for evaluating SOTA computer vision techniques \cite{resnet, swin, convnext}. Several studies have demonstrated that AT models trained on IN-1K can transfer their adversarial robustness to downstream dense-prediction tasks and achieve zero-shot adversarially robust recognition \cite{advod, advseg, zeroshot}.

Recently, two works \cite{rock, carlinipart} revisited part-based recognition models from the robustness perspective and showed the potential to enhance adversarial robustness. Li et al. \cite{rock} proposed ROCK, a part-based model that first predicts all parts represented by segmentation results, and then utilizes a judgment block to give category predictions based on the predicted parts. Similarly, Sitawarin et al. \cite{carlinipart} investigated several types of two-stage part-based models, where the first stage involves part segmentation, followed by the recognition stage with a tiny classifier. However, as mentioned before, both works were only validated on small-scale part-friendly datasets for the lack of part annotations. Sitawarin et al. \cite{carlinipart} performed classification tasks only on 11 super-categories of PIN \cite{partimagenet}. Furthermore, the proposed two-stage part-based models inevitably introduced a lot of extra parameters during inference, while the utilization of high-resolution part annotations was limited, as discussed in \cref{sec:modeldesign}. In addition, ROCK \cite{rock} contains a non-differentiable process, which could lead to a potential overestimation of adversarial robustness \cite{AthalyeC018, adaptive20}. Our work will address these issues.

\section{The PIN++ Dataset}
\label{sec:dataset}
In this section, we first present the details of how we built PIN++ to ensure high-quality part annotations, followed by the statistics of this dataset. We then give a discussion on PIN++.
\subsection{Annotation Scheme}
\label{sec:annotationscheme}

\noindent\textbf{Data source.} IN-1K \cite{imagenet}, containing about 1.3M images with category labels in the training set, is one of the most widely used datasets for general object recognition. We aimed to provide part annotations for part-based robust recognition on IN-1K, and thus PIN++ directly reuse the \textit{training} set of IN-1K. All the images conform to licensing for research purposes. However, due to cost constraints, it was impractical to provide part-level annotations for all training images in IN-1K. Instead, we decided to annotate 100 randomly selected images per category. Once part annotations are provided for all categories, modern supervised segmentation techniques \cite{maskrcnn} can be used to obtain pseudo part labels for the remaining unannotated images, as we will show in \cref{sec:MPM}. Note that PIN++ does not provide any part annotations for the validation set of IN-1K.

While PIN \cite{partimagenet} has provided some part annotations for 158 categories of IN-1K, their annotations do not fully meet our purpose and principles. To minimize annotation costs without compromising quality, we retained annotations of 90 categories from PIN as part of PIN++. The details are shown in Appendix \ref{sec:supp_reuse}. We then show the annotation design for the remaining 910 categories of IN-1K. 

\noindent\textbf{Annotation quality control.} To ensure high-quality part annotations, three points should be taken into account:

\noindent\textbf{1) Deciding which parts to annotate per category.} One of the key challenges in the part annotation task was the ambiguity of object part selection (e.g., how to annotate the parts of \textit{hammer}). PIN \cite{partimagenet} evaded this problem and focused primarily on annotating parts for animal categories (most \textit{quadrupeds} could be regarded as consisting of four part categories: \textit{head}, \textit{body}, \textit{foot}, \textit{tail}). However, the remaining categories in IN-1K were highly diverse, making it challenging to decide which parts to annotate. To ensure a scientifically grounded division of parts for each category, we initially consulted the Wikidata knowledge base\footnote{\url{https://www.wikidata.org/wiki/Wikidata:Introduction}} to obtain part vocabularies for every object category (e.g., \textit{hammer} consists of three parts according to Wikidata: \textit{handle}, \textit{striker}, and \textit{hammerhead}). Additionally, for object categories without a clear definition of parts on Wikidata, we decided which parts to annotate by asking recruited volunteers which parts they thought to be important to their cognition of the object. For example, according to the cognition of volunteers, for most \textit{quadruped} categories, annotating \textit{head}, \textit{body}, \textit{foot}, and \textit{tail} was enough, while for \textit{ram} (see \cref{fig:dataset} for one example), \textit{horn} should be annotated as an extra part. For categories that were indeed difficult to decompose into parts, e.g., \textit{flatworm}, we generalized the concept of ``parts'' and treated the foreground of an object as one part category. In this case, the part was the object itself. 

\begin{figure*}[!t]
  \centering
   \includegraphics[width=\linewidth]{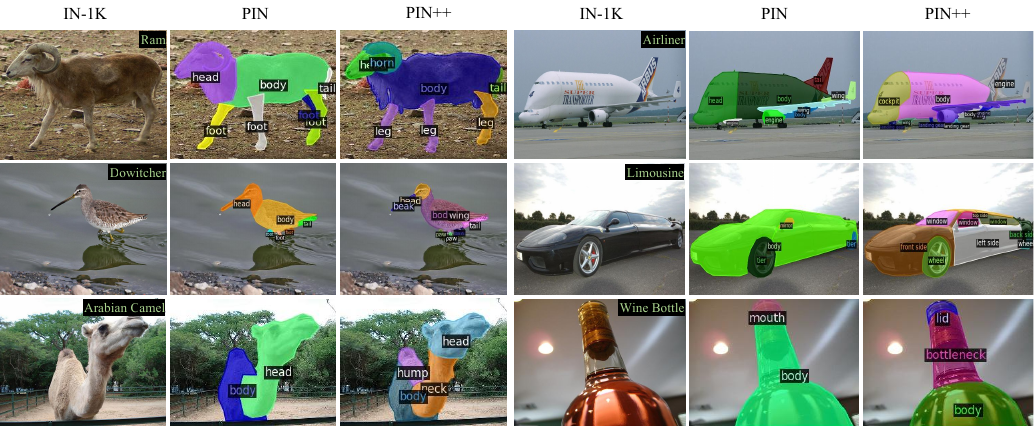}
   \caption{Visual comparison of annotations between PIN and PIN++. The object names are displayed on the top-right of the IN-1K columns. The part names are presented in the images of PIN and PIN++.}
   \label{fig:compare}
\end{figure*}

In addition to using Wikidata and the cognition of volunteers, we also used some additional rules to simplify the annotation pipeline and ensure annotation quality. Firstly, the combination of the part categories should form the complete object category. Secondly, the number of part categories was roughly set to be within the range of three to eight, except for special cases (e.g., \textit{flatworm}). When an object category was relatively simple and hard to annotate with more than three part categories, a certain part category could be added based on its function, shape, etc. For example, in the case of a \textit{maraca}, which consists of part categories \textit{head} and \textit{stick}, the \textit{stick} can be further divided into part categories \textit{joint} and \textit{handle} to better represent the gripping action for a \textit{maraca}. Thirdly, in cases where two part categories for an object overlap significantly, only the larger part category was annotated. For example, in the case of an \textit{acoustic guitar}, where the \textit{fingerboard} and \textit{strings} significantly overlap, we considered only the \textit{fingerboard} as a part category. With these rules, we could achieve high-quality part annotations for all categories in the IN-1K dataset while controlling the annotation cost. These principles ensure that all categories in IN-1K can have part annotations.

\noindent\textbf{2) Designing the part segmentation principles.} We designed several part segmentation principles to guide the annotators and ensured high-quality part segmentation annotation. Firstly, the annotated part masks were required to be combined to cover the entire object and have no overlap, unless in the case of the second principle. Secondly, for some parts that indeed should overlap in semantic concepts (e.g., \textit{horn} and \textit{head}), the annotators should annotate the inclusion relation of parts (e.g., \textit{horn} is included in \textit{head}), and then these part masks could be annotated with overlap. Thirdly, to keep the consistency of part annotations among images of the same object category, following \cite{partimagenet}, the same annotator should annotate all 100 images of this object category. The other principles are shown in Appendix \ref{sec:supp_principle}.

\noindent\textbf{3) Annotation quality inspection.} Ten randomly selected images together with the annotation visualizations (see \cref{fig:dataset}) were sent to inspectors to assess if the annotations for a specific object category met the defined principles. If the annotations for any two images failed to meet these principles, the entire category's annotations were re-annotated until the requirements were satisfied.

\subsection{Statistics}
\label{sec:statistics}
We finally annotated 100 images per category for 910 categories in IN-1K, resulting in a total of 91K images with part annotations. Taking these data and the data of 90 reused categories of PIN together, PIN++ for 1K categories was obtained. Following IN-1K, here we ensured PIN++ was class-balanced. PIN++ is split into \texttt{train/val} sets with a 9:1 ratio for each category. However, note that all annotated images are part of the IN-1K training set. Unless stated otherwise, they were all utilized for training in our experiments and the evaluation was performed on IN-1K \texttt{val} set.

In summary, PIN++ includes a total of 100K images with part annotations. These annotations cover 3,310 part categories for 1,000 object categories. A total of 406,364 part masks were annotated in PIN++. 

\begin{table}[!t]
  \centering
  \caption{The annotation density of PartImageNet++.}
    \begin{tabular}{c|cccc}
    \toprule
     Number of part masks & 1-2 & 3-4 & 5-6 & 7+  \\
     \midrule
     Proportion ($\%$) & $46.27$ & $44.87$ & $8.66$ & $0.20$ \\
     \bottomrule
    \end{tabular}   
  \label{tab:density}
\end{table}

\begin{table}[!t]
  \centering
  \caption{Comparison of PIN++ with Other Part Datasets.}
  \setlength{\tabcolsep}{1pt}
  {
    \begin{tabular}{c|cccc}
    \toprule
     Dataset  & Obj. Cate. & Part Cate. & Image & Part Mask  \\
     \midrule
     Cityscapes PP. \cite{Cityscapes-Panoptic-Parts} & $5$ & $23$ & $3.5$K & $100$K \\
     Pascal-Part \cite{PASCAL-Part} & $20$ & $193$ & $19$K & $363.5$K \\
     ADE20K$^\star$ \cite{ade20k}  & $80$ & $566$ & $12.6$K & $193.2$K \\
     PACO \cite{paco} & $75$ & $456$ & $76.7$K& $\mathbf{641.4}$K \\
     PIN \cite{partimagenet} & $158$ & $609$ & $24$K & $112$K \\
     \rowcolor{lightgray} PIN++ & $\mathbf{1000}$ & $\mathbf{3308}$ & $\mathbf{100}$K & $406.4$K  \\
     \bottomrule
    \end{tabular}
    \begin{tablenotes}
            \footnotesize
            \item $^\star$: ADE20K is severely category-unbalanced and here we show statistics for categories with more than 10 annotated part masks.
    \end{tablenotes}
    }
  \label{tab:comparedataset}
\end{table}

\noindent\textbf{Annotation cost and density.} 
PIN++ was annotated by 50 annotators who collectively invested approximately 8,000 hours in the annotation process. In addition, 10 volunteers were involved in determining the parts to be annotated and 5 inspectors were involved in the annotation quality inspection. During this process, 37,505 low-quality images were discarded, with each discarded image reviewed by two inspectors. With such efforts, PIN++ possesses a substantial number of part mask annotations, resulting in 406,364 mask annotations. We calculated the proportion of part mask annotations per image, which we called the density of annotations, as shown in \cref{tab:density}. Notably, over half of the images contain 3-6 part mask annotations. Given the extensive number of images in PIN++, it offers abundant training supervision for various part-related visual tasks.

In addition, PIN++ provides inclusion relations for annotated part masks that exhibit overlap. These relations are organized in a dictionary format, categorized by each object category, resulting in a total of 201 object categories with corresponding 317 inclusion relations. These inclusion relations serve as valuable resources for researchers to understand the hierarchical structure of the annotated part categories.


\begin{figure*}[!tb]
  \centering
   \includegraphics[width=0.9\linewidth]{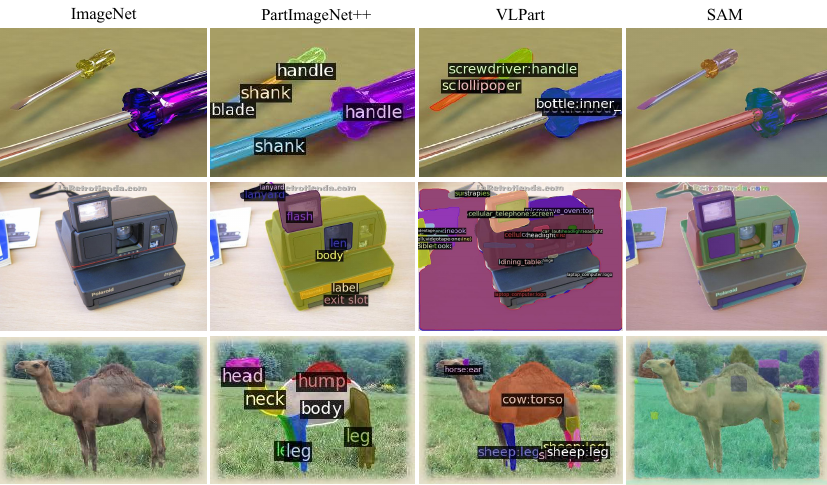}
   \caption{Comparison between part segmentation results of different methods and PIN++ annotations. Without training on PIN++, VLPart and SAM fail to segment objects into specific parts with accurate semantics.}
   \label{fig:datasetcompare}
\end{figure*}

\subsection{Comparison with Other Part Datasets} 
We compared PIN++ with publicly available part datasets, as shown in \cref{tab:comparedataset}. PIN++ outperforms previous datasets in terms of the number of object and part categories, as well as the total number of images with annotated part masks. It also demonstrates the competitive number of part masks compared to the recent PACO dataset \cite{paco}. Notably, PIN++ offers part annotations for a diverse range of objects, including creatures, artifacts, rigid objects, and nonrigid objects, unlike PACO or PIN, which are primarily focused on common tools or animals, respectively. As shown in \cref{sec:statistics}, the annotation quality of PIN++ exhibits an advantage over that of PIN. Moreover, PIN++ is category-balanced and guarantees that each image contains only one foreground category. All these features build the unique advantage of PIN++ for studying object recognition tasks. 

\cref{fig:compare} further shows some visual annotations from the 68 object categories annotated in both PIN and PIN++ (see Appendix \ref{sec:supp_reuse}). The comparison indicates that PIN++ provides superior annotation quality compared to PIN. PIN++ exhibits finer-grained part segmentation and more accurate part names and effectively captures the distinctive characteristics of each object category.

\subsection{Discussion on PIN++}
Recent advancements in segmentation included CLIP-based open-vocabulary segmentation methods \cite{glip, opendet, vlpart} and the SAM model \cite{sam}. We compared these techniques with the part annotations of PIN++. We used the largest version of SAM, and for open-vocabulary segmentation, we employed the VLPart model \cite{vlpart}, which was trained with part annotations from several part datasets including Pascal-Part \cite{PASCAL-Part}, PIN \cite{partimagenet}, and PACO \cite{paco}. Visual comparisons are presented in \cref{fig:datasetcompare}. We can see that VLPart only provided roughly reasonable semantics for part categories it had been trained on. It failed to accurately identify the semantics of parts from unseen objects. On the other hand, SAM seemed to perform well in segmenting certain parts based on the object's edges, but it struggled when the part edges were less distinguishable, as shown in the segmentation results for \textit{camel} (the third row in \cref{fig:datasetcompare}). In addition, note that all masks segmented by SAM were category-agnostic.

The above results highlight the challenges faced by existing techniques in achieving accurate part segmentation for diverse objects without training on PIN++. Additionally, current text-to-image generative models \cite{stablediffusion} also struggle with understanding part semantics \cite{diffusion}. However, supervised training on PIN++ yielded good part segmentation results, as shown in \cref{fig:pseudolabel}.
Therefore, besides its usage in robust object recognition (as demonstrated next), we believe that PIN++ and its extensive collections of 3308 part categories (vocabularies) have the potential to enhance general part-related visual understanding tasks.

\section{Part-Based Robust Object Recognition}
\label{sec:MPM}

\begin{figure*}[!t]
  \centering
   \includegraphics[width=0.95\linewidth]{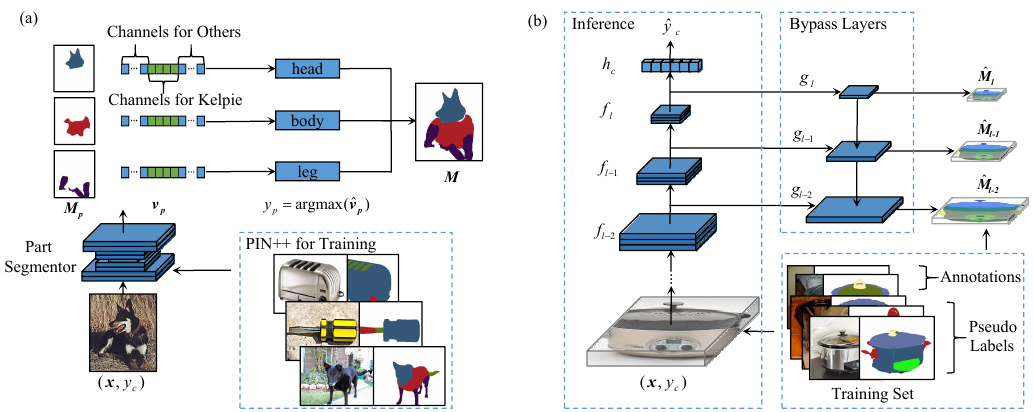}
   \caption{An overview of the generation of pseudo-labels and the structure of MPM. (a) A part segmentation model trained on PIN++ and used to obtain pseudo-part labels for unannotated images. (b) MPM adds several auxiliary bypass layers to the vanilla recognition model for part segmentation supervision. MPM is trained by part annotations together with the pseudo part labels. During inference, the auxiliary layers are dropped, and the vanilla recognition model gives the final object category prediction. }
   \label{fig:structure}
\end{figure*}

With the annotations of PIN++, we develop part-based methods directly on the standard IN-1K. Our part-based methods can be divided into two steps. We first obtain pseudo part labels for the remaining unlabeled images of IN-1K and then train the MPM with these part annotations and pseudo labels. The pipeline of our part-based methods is illustrated in \cref{fig:structure}.

\subsection{Pseudo Part Label Generation}
\label{pseudo-generation}
\noindent\textbf{Part segmentation model.}
Arbitrary instance segmentation models \cite{maskrcnn, instanceasquery} can be leveraged to perform part segmentation and generate pseudo part labels when trained with the part annotations from PIN++. Following Sun et al. \cite{vlpart}, we use the classical Mask R-CNN \cite{maskrcnn} model with a vision transformer Swin-B \cite{swin} as the backbone network. In training the Mask R-CNN, we consider the masks for each part category as masks for individual object categories directly. During inference for obtaining pseudo part labels, we incorporate a simple post-processing operation alongside the regular inference procedure. 

\noindent\textbf{Post-processing.}
Given an image $\mathbf{x} \in \mathbb{R}^{3 \times H \times W}$ with object category $y_c$, where $H \times W$ represents the resolution, the regular inference procedure of the trained Mask R-CNN produces pseudo part labels in the form of $\{(\mathbf{M_p}, \mathbf{v_p})\}$, where $\mathbf{M_p} \in \{0, 1\} ^ {H \times W}$ denotes the binary part mask, $\mathbf{v_p} \in [0, 1] ^ {K}$ denotes the output probability of part categories, and $K$ denotes the number of part categories (3308 in our case). Note that the images that need inference are all in the training set of IN-1K. Their object category $y_c$ is not utilized in the regular inference procedure. To leverage it, we apply a Category Filter (CF) operation to obtain $\mathbf{\hat{v}_p}$, which sets the probabilities of unrelated part categories in $\mathbf{v_p}$ to zero (e.g., for \textit{cat}, except for \textit{cat:head}, \textit{cat:body}, etc, other part categories such as \textit{radio:button} and \textit{bird:head} are all excluded). The pseudo part category for $\mathbf{M_p}$ is then computed as $y_p = \argmax \mathbf{\hat{v}_p}$.


\subsection{Multi-scale Part-supervised Recognition Model}
\label{sec:modeldesign}
\begin{figure*}[!t]
  \centering
   \includegraphics[width=\linewidth]{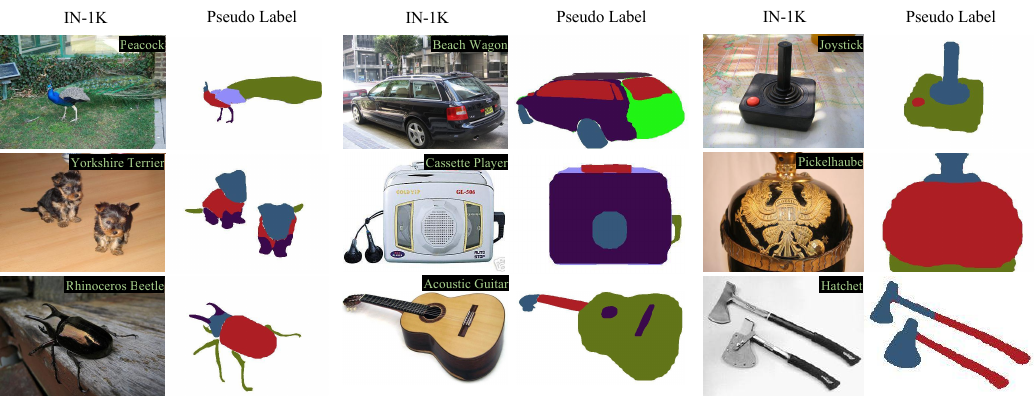}
   \caption{Visualization of pseudo part labels generated by a Mask R-CNN trained on PIN++. The object names are shown on the top-right of each image. The part names are hidden here for clarity.}
   \label{fig:pseudolabel}
\end{figure*}

\noindent\textbf{Data preparation.}
After obtaining the pseudo labels for all remaining training images of IN-1K, we simply treat the pseudo labels and real part annotations equally because we find that the quality of the pseudo labels is roughly satisfactory, as shown in \cref{fig:pseudolabel}. We provide quantitative results for pseudo-label quality later. The part labels $\{(\mathbf{M_p}, \mathbf{y_p})\}$ for an image $\mathbf{x}$ are then converted to a single composite segmentation mask $\mathbf{M} \in \{0, 1\} ^ {H \times W \times (K + 1)}$, where the extra one channel indicates the background. Finally, the data samples for training part-based models are $\{(\mathbf{x}, y_c, \mathbf{M})\}$.

\noindent\textbf{Model design.} 
Both previous works \cite{rock, carlinipart} build part-based models in a two-stage way: an effective image segmenter $\mathcal{F}_{\mathrm{seg}}: \mathbb{R} ^ {3 \times H \times W}\rightarrow \mathbb{R} ^ {(K+1) \times h \times w}$ and an extra module $\mathcal{F}_{\mathrm{cls}}: \mathbb{R} ^ {(K+1) \times h \times w}\rightarrow\mathbb{R}^{C}$ for classification, where $h \times w$ denotes the output resolution and $C$ denotes the number of object categories. $\mathcal{F}_{\mathrm{seg}}$ is a backbone network (e.g., a ResNet-50 without classification head). Most popular backbones \cite{resnet, convnext, swin} are hierarchically composed of several blocks with down-sampling layers and thus they can be written as $\mathcal{F}_{\mathrm{seg}}:= f_l \circ \cdots \circ f_1$, where $l$ denotes the number of down-sampling layers and $f_{i(1 \leq i \leq l)}$ denotes one block with the $i$th down-sampling layer. $\mathcal{F}_{\mathrm{cls}}$ can be a tiny classifier \cite{carlinipart} or a non-differentiable judgment block \cite{rock} and its input is the part segmentation results $\hat{\mathbf{M}} = \mathcal{F}_{\mathrm{seg}}(\mathbf{x})$. The overall model for recognition is  $\mathcal{F} := \mathcal{F}_{\mathrm{cls}}\circ\mathcal{F}_{\mathrm{seg}}$. However, as $K$ is significantly larger than $C$, using $\hat{M}$ as the intermediate results for recognition inevitably introduces extra parameters and computation during inference. This situation gets worse with an increased number of part categories. Moreover, these methods solely employ part supervision on the output of $f_l$. Without expanding $\mathcal{F}_{\mathrm{seg}}$, part annotations have to be down-sampled to match the output resolution of $f_l$, potentially limiting the utilization of high-resolution part annotations. 

Instead of these two-stage part-based models, MPM is expected to learn a robust intermediate representation by adding multi-scale bypass layers to the vanilla recognition network. In MPM, the model used for object recognition is the vanilla backbone network directly: $\mathcal{F} := h_c \circ f_l \circ \cdots \circ f_1$, where $h_c$ denotes a vanilla classification head (usually a linear layer). MPM utilizes the part annotations to supervise the intermediate features of $\mathcal{F}$ by several bypass layers $g_i$: $\hat{\mathbf{M}_i} = g_i \circ f_i \circ \cdots \circ f_1(\mathbf{x})$, where $i \leq l$.

In this way, the intermediate features of $\mathcal{F}$ can be seen as implicit representations of part segmentation results when $g_i$ is simple enough. Note that here we use part supervision not only on the output of $f_l$, but also $f_{i(i<l)}$. As the output of lower layers has a larger resolution, it can be supervised by part annotations with higher resolution. But the outputs of $f_{i(i<l)}$ are relatively low-level, which may not be enough to obtain high-level part segmentation results themselves.

To balance this, only the outputs of the last three blocks $f_{i(l - 2 \leq i \leq l)}$ are supervised by part mask $\mathbf{M}$ (e.g., for a ResNet-50 with input size $224 \times 224$, the intermediate features of $7 \times 7$, $14 \times 14$, and $28 \times 28$ are supervised by corresponding down-sampled $\mathbf{M}$, respectively). 

In addition, following the principle of FPN \cite{fpn}, some top-down layers are used to augment the low-level features with high-level features. Note that distinct from the conventional FPN that generally has comparable parameters with the backbone network \cite{fpn}, these bypass layers are extremely lightweight as our goal is to improve the recognition accuracy rather than the quality of part segmentation results. \cref{fig:structure}(b) shows the overall structure of MPM. During inference, these auxiliary layers are dropped, and the vanilla recognition model gives the final object category prediction. 
The overall loss for training MPM is: $L = L_{\mathrm{cls}} + \lambda \cdot L_{\mathrm{seg}}$, where $L_{\mathrm{cls}}$ denotes the vanilla loss for classification, $L_{\mathrm{seg}}$ denotes the loss for part segmentation, and $\lambda$ is a hyper-parameter. We compute $L_{\mathrm{seg}}$ as the average of losses on three part segmentation results of different resolutions, as mentioned before. Note that when performing adversarial training, adversarial examples are generated by $\mathbf{x}^{\star} = \argmax_{\mathbf{x}^{\star}: ||\mathbf{x}^{\star} - \mathbf{x}||_p \leq \epsilon} L_{\mathrm{cls}}(\mathcal{F}(\mathbf{x}^{\star}), y)$, while MPM is always trained with $L$.


\subsection{Experimental Setup on MPM}
\label{sec:setting}

\noindent\textbf{Training setup.} Unless specified otherwise, following previous works \cite{bairobust, debenedetti}, we performed AT with $l_\infty$ bound $\epsilon = 4/255$ on IN-1K. The inner optimization for obtaining adversarial examples used PGD \cite{at} with iterative steps $t = 2$ and the step size is set to be $s = 2 * \epsilon / t$. The input resolution used $224 \times 224$. We used the vanilla version of the AT \cite{at} and did not incorporate other variants such as TRADES \cite{trades} or AWP \cite{awp}. These variants have shown effectiveness mainly on small-scale datasets, making their generalization to IN-1K non-trivial. Following previous part-based works \cite{rock, carlinipart}, we used the classical ResNet-50 \cite{resnet} network with 25.6M parameters as the baseline model. We note that recent works \cite{bairobust, debenedetti} used a GELU \cite{gelu} activation to replace the original RELU \cite{relu} activation of ResNet-50 for boosting robustness when performing AT. Here we also used GELU, denoted as ResNet-50-Gelu. In addition, we improved the training recipe of ResNet-50 to build a strong baseline. Other details on the training recipe are provided in Appendix \ref{sec:supp_recipe}. It should be emphasized that MPM employs bypass layers with approximately 4.5 million parameters during the training process. However, these bypass layers are disregarded during the inference stage.


\noindent\textbf{Evaluation setup.}
For all experiments in the work, unless specified otherwise, we evaluated the adversarial robustness with AutoAttack \cite{autoattack}, which is a combination of various attack methods and is generally recognized as a reliable evaluation \cite{robustbench}. The evaluation was performed at a resolution $224 \times 224$ on randomly selected 10K images of the 50K IN-1K \texttt{val} set, 10 images for each category. This ensured that the 95\% confidence intervals of the reported average adversarial robustness (see \cref{tab:reeval}) were less than $\pm0.5\%$. Note that, unlike previous works \cite{bairobust, debenedetti}, we did not use the 5K images selected by RobustBench \cite{robustbench} as we found that they were category-unbalanced, containing three categories even without any image for validation.

We expect the AT models with $l_\infty$ bound $\epsilon = 4/255$ to be adversarially robust not only on seen threats (i.e., $l_\infty$ attack with $\epsilon = 4/255$) but also on unseen attack threats (e.g., $l_\infty$ attack with larger $\epsilon$, $l_1$ and $l_2$ attacks). Thus, we evaluated the models on three attack threats: $l_\infty$, $l_1$, and $l_2$, with the bounds $\epsilon_\infty = 4/255$, $\epsilon_1 = 75$, $\epsilon_2 = 2$, respectively. For $l_\infty$ attack, we extra evaluated the models with $\epsilon_\infty = 8/255$ to mimic a stronger threat, denoted as $l_\infty(\times 2)$.

\subsection{Experimental Results on MPM}
\noindent\textbf{Robustness against Adversarial Attacks.}
Utilizing the aforementioned configuration, we commenced the training process with the standard ResNet-50-Gelu serving as our performance benchmark. For a fair comparison, MPM adhered to the same training protocol as the baseline model. Additionally, an $L_\mathrm{seg}$ term was incorporated to include part-based supervision, with $\lambda$ assigned a value of 1. Furthermore, we compared MPM's performance with that of all publicly available ResNet-50 models that had undergone adversarial training on IN-1K in recent years. The tabulation of recognition accuracies under various attack scenarios is presented in \cref{tab:reeval}. Evidently, our baseline model displayed a robust performance, outperforming all comparative benchmarks by an average of 2.2\% in terms of robustness metrics (17.1\% as opposed to 19.3\%). MPM demonstrated a significant improvement in adversarial robustness over the standard baseline across the entire range of attack vectors. Moreover, MPM also enhances the accuracy of clean image recognition when compared to our robust baseline model. To the best of our knowledge, these results represent the state of the art in adversarial robustness for ResNet-50 on the IN-1K dataset. 

\begin{table*}[!t]
  \centering
  \caption{Recognition accuracies ($\%$) of methods under different attack threats on IN-1K, following the evaluation setup in \cref{sec:setting}. \textit{Average} denotes the average accuracies in four attack settings. We highlight the best results in each column.}
  \small
   \setlength{\tabcolsep}{4pt}
   {
    \begin{tabular}{cc|ccccc|c}
        \toprule
	  \multirow{2}*{Architecture}	& \multirow{2}*{Method} & \multicolumn{6}{c}{Adversarial Train with respect to $l_\infty$ ($\epsilon = 4/255$)} \\
	  	  &  &  Clean & $l_\infty$ & $l_\infty (\times 2)$ & $l_1$ & $l_2$ &  Average  \\
    \midrule
	  
	  ResNet-50 & Salman et al. \cite{transfer} & $63.9$ & $35.9$ & $13.2$ & $2.0$ & $13.7$ & $16.2$ \\
	  ResNet-50 & Mao et al. \cite{easyrobust} & $64.7$ & $34.3$ & $11.4$ & $2.3$ & $16.4$ & $16.1$ \\
        ResNet-50-Gelu & Bai et al. \cite{bairobust} & $\mathbf{67.8}$ & $36.6$ & $10.8$ & $3.0$ & $16.8$ & $16.8$ \\
	  ResNet-50-Gelu & Debenedetti et al. \cite{debenedetti} & $66.8$ & $35.6$ & $12.8$ & $3.6$ & $16.5$ & $17.1$ \\
      ResNet-50 & Liu et al. \cite{liu} & $65.5$ & $32.6$ & $9.5$ & $3.3$ & $17.9$ & $15.8$ \\
	  \rowcolor{lightgray} ResNet-50-Gelu & ours (vanilla) & $67.1$ & $38.1$ & $12.6$ & $5.0$ & $21.6$ & $19.3$ \\
	  \rowcolor{lightgray} ResNet-50-Gelu & ours (MPM) & $\mathbf{67.8}$ & $\mathbf{39.1}$ & $\mathbf{13.6}$ & $\mathbf{6.2}$ & $\mathbf{24.3}$ & $\mathbf{20.8}$ \\
        \bottomrule	  
    \end{tabular}
    }
  \label{tab:reeval}%
\end{table*}

\begin{table}[!t]
  \centering
  \caption{Recognition accuracies (\%) of MPM and ROCK \cite{rock} on 125 categories of PIN. $l_\infty$ and $l_\infty (\times 2)$ indicates $l_\infty$ attacks with the bounds $\epsilon_\infty = 4/255$ and $\epsilon_\infty = 8/255$, respectively.}
  \small
  {
    \begin{tabular}{c|ccc}
    \toprule
     Method  &  Clean & $l_\infty$ & $l_\infty (\times 2)$  \\
     \midrule
     Li et al. \cite{rock} & $54.5$ &  $34.2$ & $17.3$ \\
     ours (MPM) &   $\mathbf{70.0}$ & $\mathbf{42.3}$ & $\mathbf{18.9}$\\
     
     \bottomrule
    \end{tabular}   
    }
  \label{tab:class125}
\end{table}

\begin{table}[!t]
  \centering
  \caption{Recognition accuracies (\%) of MPM and the part-based model proposed by Sitawarin et al. \cite{carlinipart} on 11 super-categories of PIN.}
  \small
  {
    \begin{tabular}{c|ccc}
    \toprule
     Method  &  Clean &$l_\infty$ & $l_\infty (\times 2)$  \\
     \midrule
     Sitawarin et al. \cite{carlinipart} & $85.6$ & - & $39.4$ \\
     ours (MPM) & $\mathbf{91.4}$ & $72.3$ & $\mathbf{43.7}$ \\
     
     \bottomrule
    \end{tabular}   
    }
  \label{tab:class11}
\end{table}

Furthermore, we undertake a comparison between the recognition accuracies of MPM trained on PIN++ and previous part-based models \cite{rock, carlinipart} trained on PIN. In accordance with previous works, adversarial training (AT) for MPM is carried out with an $l_{\infty}$ bound of $\epsilon = 8/255$. The training setup delineated in \cref{sec:setting} is adhered to, guaranteeing consistency across experiments.

While the MPM is capable of performing classification on 1000 categories of PIN++, previous studies \cite{rock, carlinipart} report results on subsets of PIN. In order to be in line with these studies, we mask out the other category channel of MPM and present the results on the corresponding subsets. Specifically, the results on 125 categories of PIN are reported in \cref{tab:class125}, following the setting of Li et al. as mentioned in \cite{rock}. Additionally, the results on 11 super-categories of PIN are reported in \cref{tab:class11}, in accordance with the setting of Sitawarin et al. \cite{carlinipart}.

The results attained through MPM trained on PIN++ exhibit substantial enhancements compared to previous outcomes achieved on PIN. Nevertheless, it should be noted that caution must be exercised when interpreting these results, as the data employed for training are different and not directly comparable.

\begin{table}[!t]
  \centering
  \caption{Segmentation accuracies (AP) of Mask R-CNN with and without CF, and recognition accuracies (\%) of MPM on IN-1K trained with various pseudo labels.}
  \small
  \setlength{\tabcolsep}{2pt}
  {
    \begin{tabular}{cc|cc|ccc}
    \toprule

     Pseudo Label & CF & AP & AP$_{50}$ & Clean & $l_\infty$  & $l_\infty (\times 2)$ \\
     \midrule
      $$ & $$  & - & - & $65.5$ & $33.0$ & $9.4$\\
       $\checkmark$ & $ $ &  $37.2$ & $58.6$ & $\mathbf{67.8}$ & $38.8$ & $13.1$\\
       $\checkmark$ & $\checkmark$  & $\mathbf{40.7}$ & $\mathbf{64.4}$ & $\mathbf{67.8}$ & $\mathbf{39.1}$ & $\mathbf{13.6}$\\
     \bottomrule
    \end{tabular}  
    }
  \label{tab:pl_quality}
\end{table}

\begin{table}[!t]
  \centering
  \caption{Recognition accuracies (\%) of different part-based models on IN-1K. The number of parameters during inference is listed.}
  \small
  \setlength{\tabcolsep}{2pt}
  {
    \begin{tabular}{c|c|ccc}
    \toprule
     Method & Param. &  Clean &$l_\infty$ & $l_\infty (\times 2)$  \\
     \midrule
     Two-stage & $60.9$M &  $67.3$ & $38.7$ & $13.0$\\
     \hline
     SV1 & \multirow{4}*{$25.6$M} &  $67.3$ & $36.9$ & $8.8$\\
     SV2 w/ TD &   & $67.6$ & $37.1$ & $11.7$\\
     MPM (SV3 w/ TD)&  &  $\mathbf{67.8}$ & $\mathbf{39.1}$ & $\mathbf{13.6}$\\
     SV3 w/o TD & & $67.4$ & $36.9$ & $11.7$ \\
     
     \bottomrule
    \end{tabular}   
    }
  \label{tab:diff_model_structure}
\end{table}


\noindent\textbf{Models versus Humans.} 
The part-based models are motivated by the human recognition process. We were interested in whether the robustness of MPM aligns more with human cognition. To investigate this, we employed the evaluation method introduced by Geirhos et al. \cite{modelvshuman}, which involves comparing the models' decisions on different distorted images with the actual judgments made by human observers. 

Three metrics quantify how closely model predictions are aligned with the decisions of humans \cite{modelvshuman}: \textit{Accuracy Difference} measures the average accuracy disparity between a model and human observers across different image distortions. \textit{Observed Consistency} quantifies the agreement between model and human decisions on the same samples. It measures the fraction of samples where both the model and humans make the same decision, regardless of whether it is correct or not. \textit{Error Consistency} is the key metric, which measures the shared mistakes between the model and humans, indicating consistency in error patterns. When the accuracies of both models and humans are high, they can achieve high observed consistency with very different decision strategies, while error consistency can still track whether there is above-chance consistency in decision-making. 

The results shown in \cref{tab:modelvshuman} demonstrate that integrating MPM and adversarial training significantly improves alignment with human cognition. The most substantial gains occur when AT is applied: AT alone reduces the accuracy difference to 0.069 and elevates error consistency to 0.252, suggesting that adversarial robustness plays a critical role in mimicking human error patterns. Combining both part-based modeling and AT achieves the best overall alignment, with the lowest accuracy difference (0.069), highest observed consistency (0.679), and highest error consistency (0.261). These results confirm that MPM, particularly when augmented with adversarial training, aligns more closely with human cognitive strategies, as evidenced by reduced performance gaps and increased shared decision patterns—including error tendencies. This supports the hypothesis that part-based hierarchical reasoning, coupled with robustness to distortions, reflects principles inherent to human visual recognition.

\begin{table}[t]
  \centering
  \caption{Comparison of recognition results between different models and human decisions \cite{modelvshuman}. The direction of the arrows indicates better alignment between the models and humans.}
  \setlength{\tabcolsep}{2pt}
  {
    \begin{tabular}{cc|ccc}
    \toprule
     Part & AT & Acc. Diff.$\downarrow$ & Obs. Consistency$\uparrow$ & Error Consistency$\uparrow$  \\
     \midrule
     $ $ & $ $ & $0.083$ & $0.665$ & $0.195$  \\
     $\checkmark$ & $ $ & $\mathbf{0.081}$ & $\mathbf{0.668}$ & $\mathbf{0.204}$ \\
     \hline
     $ $ & $\checkmark$ & $\mathbf{0.069}$ & $0.676$ & $0.252$ \\
     $\checkmark$ & $\checkmark$ & $\mathbf{0.069}$ & $\mathbf{0.679}$ & $\mathbf{0.261}$ \\
     \bottomrule
    \end{tabular}
    }
  \label{tab:modelvshuman}
\end{table}

\noindent\textbf{Robustness  on Corruptions and OOD Datasets.} 
Appendix \ref{sec:supp_ood} shows that MPM also exhibits improved robustness on out-of-distribution shifts. MPM outperforms non-part models across nearly all corruption types (e.g., noise, blur, weather distortions), with results averaged over five severity levels, regardless of whether AT is applied. MPM shows improved performance on four OOD datasets—ImageNet-A-Plus, ImageNet-Sketch, Stylized ImageNet (SIN), and a distortion dataset (DIN) -- highlighting its adaptability to realistic distribution shifts.

\noindent\textbf{Other Advantages of MPM.} 
In addition, MPM enhances both the clean accuracy and adversarial robustness in object detection.  
Li et al. \cite{advod} show that adversarially trained models on the large-scale IN-1K can be utilized to initialize the backbone of downstream networks (e.g., Faster R-CNN \cite{fasterrcnn}). This approach together with downstream AT enables the transfer of adversarial robustness to downstream tasks. We conducted similar experiments by using the checkpoints of the vanilla baseline and MPM. The results shown in Appendix \ref{sec:supp_downstream} indicate that MPM enhances both the clean accuracy and adversarial robustness in object detection. 

\noindent\textbf{Ablation Study on MPM.} 
We first investigate the influence of pseudo-label quality on the adversarial robustness of MPM. In \cref{pseudo-generation}, we illustrate the use of CF during generating pseudo labels. Here we remove the CF operator and regenerate pseudo labels for images without part annotations. The left half of \cref{tab:pl_quality} (the 2nd and 3rd rows) shows the segmentation accuracies of models with and without CF (here the models were supervised with the \texttt{train} set of PIN++, and the AP were calculated on \texttt{val}). It is seen that training with PIN++ achieved good part segmentation results, indicating its potential for enhancing part-related tasks. CF further significantly improves AP. 

We then trained MPM using both pseudo labels and real part annotations, as well as solely using real part annotations. The results, displayed in \cref{tab:pl_quality} (the 1st row), indicate that training MPM solely with real part annotations is ineffective, potentially compromising the model's performance. We hypothesize it could be caused by overfitting on the images with part annotations. Besides, the inclusion of better-quality pseudo labels can improve the robustness of our part-based model, underscoring the importance of high-quality pseudo labels.

We further conducted ablation experiments on the structure of the part-based models. Specifically, we compared MPM with the two-stage part-based model designed by Sitawarin et al. \cite{carlinipart}. In addition, we performed ablations on the number of part supervisions. MPM incorporates part supervision with three resolutions, referred to as SV3. Here we compared it with SV1 and SV2, which indicate part-based models solely supervised by $f_l$ and both $f_l$ and $f_{l-1}$, respectively. Besides these, we performed an ablation by removing all top-down (TD) connections of MPM. The results are shown in \cref{tab:diff_model_structure}. We can see that although the two-stage model needs about twice the parameters during inference, its robustness is inferior to the MPM. All the designs of MPM, including utilizing higher resolution part annotations and top-down connections, contribute to the robustness. 

Further ablation studies underscore the importance of fine-grained part annotations over object segmentation annotations. Results conducted on Tiny-ImageNet \cite{tinyimagenet} emphasize the need to explore adversarial robustness in high-resolution images. Additionally, part-based models demonstrate minimal sensitivity to the parameter $\lambda$. For detailed information regarding these ablation experiments, refer to Appendix \ref{sec:supp_ablation} for details.

\section{PIN++ for part and object segmentation}
\label{sec:ExperimentandBenchmark}
In this section, we conduct comprehensive experimental evaluations on the PIN++ dataset, focusing on two critical tasks: semantic part segmentation, object segmentation. Our results demonstrate that incorporating part annotations significantly improves model performance across these tasks, underscoring their utility in advancing segmentation accuracy and generalization in data-scarce scenarios. 

\subsection{Part Segmentation}
\label{sec:partsetting}
\noindent\textbf{Benchmark design.} For semantic part segmentation, we adopted a rigorous methodology for model training and evaluation. We implemented two Mask R-CNN models \cite{maskrcnn}, a widely recognized architecture for segmentation tasks due to its dual-output design. Mask R-CNN extends Faster R-CNN by introducing a parallel mask prediction branch alongside the existing bounding box regression and classification heads, enabling pixel-level segmentation.
We also incorporated ViT-Det \cite{vitdet}, a Vision Transformer variant optimized for detection and segmentation. Unlike standard ViT architectures, ViT-Det employs hierarchical feature fusion and task-specific decoders, making it particularly suited for fine-grained segmentation tasks. 

Inspired by \cite{paco}, we also explored the cascade versions \cite{cascade} of these models. 
During the training process, we used a federated loss function \cite{fedloss}. To further enhance the training process, we applied the LSJ \cite{lsj} augmentation. This method helps the model learn more robust features by exposing it to a wide range of image sizes and aspect ratios. 
All of the aforementioned models were trained with feature pyramid networks (FPNs) \cite{fpn}. For the experimental setup in the part segmentation task, we randomly sampled images from the PIN++ dataset to divide it into training, validation, and test sets following a ratio of 8: 1: 1. 

For part segmentation evaluation, we adopt two standard metrics from object detection literature: AP$_{50}$ and AP. Both metrics are computed for part mask and part bounding box predictions separately. Specifically, AP$_{50}$ is derived by first calculating the average precision (AP) at an intersection-over-union (IoU) threshold of 0.5 for each category, then averaging these values across all categories. The AP metric extends this by averaging precision scores across multiple IoU thresholds ranging from 0.5 to 0.95 with a step size of 0.05. 

We did not compare models trained on PIN with those trained on PIN++ as the number of categories in PIN++ significantly exceeds that of PIN. Specifically, PIN++ comprises 3,308 part categories, whereas PIN contains only 609. Furthermore, the two datasets exhibit substantially different distributions. PIN++ encompasses a highly diverse range of object categories, such as creatures, artifacts, rigid objects, and non-rigid objects, while PIN focuses solely on animals. Models generally find tasks more challenging on PIN++ than on PIN. Thus, direct comparison between the two datasets would be inappropriate.

\begin{table}[!t]
  \centering
  \caption{Part segmentation performance (AP / AP$_{50}$) of different backbones and architectures on part mask and bounding box predictions.}
  \small
 \setlength{\tabcolsep}{3pt}
  {
    \begin{tabular}{c|cc|cc}
    \toprule
     \multirow{2}*{Model}  & \multicolumn{2}{c|}{Mask} & \multicolumn{2}{c}{Box}  \\ 
             & AP  & \multicolumn{1}{c|}{AP$_{50}$}
             & AP  & \multicolumn{1}{c}{AP$_{50}$} \\
    \midrule
     ResNet-50 FPN &$28.2$&$45.8$&$30.0$&$50.6$\\
     ResNet-50 FPN+Cascade &$28.7$&$46.4$&$31.2$&$51.0$\\
     ResNet-101 FPN &$30.3$&$46.9$&$31.6$&$53.2$\\
     ResNet-101 FPN+Cascade &$30.3$&$46.8$&$31.8$&$\mathbf{53.5}$\\ 
     ViT-s FPN &$30.5$&$47.1$&$31.5$&$52.9$\\
     ViT-s FPN+Cascade &$\mathbf{32.0}$&$\mathbf{47.4}$&$\mathbf{32.2}$&${53.1}$\\
     \bottomrule
    \end{tabular}   
    }
  \label{tab:partseg}
\end{table}

\noindent\textbf{Results.} As detailed in \cref{sec:partsetting}, we trained Mask R-CNN models using ResNet-50 and ResNet-101 backbones, as well as a ViT-det small model and their corresponding Cascade variants. \cref{tab:partseg} offers a comprehensive overview of these models' performance on the PIN++ dataset for the part segmentation task. The results confirm our expectation that the cascade variants generally outperform their non-cascade counterparts. This is clearly seen in both the Mask and Box columns of the table, which reflect segmentation performance for instance segmentation and bounding box detection, respectively. For example, the ViT-small model with cascade demonstrates a notable improvement in AP for both Mask and Box tasks compared to the base ViT-small model without cascade. 

In addition, the table shows that the ResNet-101 backbone had a slight edge over the ResNet-50 backbone in terms of AP$_{50}$ for both Mask and Box tasks. Despite its smaller size, the ViT-small model exhibited competitive performance, underscoring the potential of transformer-based architectures in computer vision. These results offer useful guidance for researchers who are interested in part segmentation for various applications. 

\subsection{Object Segmentation}
\noindent\textbf{Benchmark design.} It is important to note that taking the union set of all part-level masks for an object gives us the object-level mask. This allows us to establish a benchmark for object segmentation. 

For the object segmentation task, we used the same experimental settings as for the part segmentation task. We trained two Mask R-CNN models and ViT-Det, together with their cascade versions. The aim of this training was to explore whether adding part-level information would improve the models' performance in object segmentation. For this purpose, we used two supervised approaches: object-only and object+part (as detailed in \cref{sec:partsetting}). The dataset split ratios for the object segmentation task were the same as those for the part segmentation task. 

We reported the AP$_{50}$ and AP on the test set. For models with relatively small capacity, the object+part supervision offered a slight improvement over the object-only supervision. However, for models with larger capacity, the object+part supervision significantly boosted model performance. This suggests that incorporating part-level information can be particularly beneficial for more complex and powerful models, potentially leading to more accurate object segmentation results. 

\begin{table}[t]
  \centering
  \caption{Object segmentation performance (AP / AP$_{50}$) across different models and training data configurations (object only vs. object+part) for both mask and bounding box predictions. } 
  \setlength{\tabcolsep}{1pt} 
  \begin{tabular}{c|c|cc|cc} 
    \toprule
    \multirow{2}{*}{Model} & \multirow{2}{*}{Train Set} & \multicolumn{2}{c|}{Mask} & \multicolumn{2}{c}{Box} \\ 
    &  & AP & AP$_{50}$ & AP & AP$_{50}$ \\
    \midrule
    {ResNet-50 } & \multirow{6}*{object} & $52.7$ & $73.0$ & $55.0$ & $75.2$ \\ 
    {ResNet-50 +cascade} &  & $53.7$ & $73.6$ & $55.2$ & $75.3$ \\ 
    {ResNet-101 } &  & $55.1$ & $74.5$ & $58.3$ & $76.6$ \\
    {ResNet-101 +cascade} &  & $55.2$ & $\mathbf{74.7}$ & $58.4$ & $76.6$ \\
    {ViT-s} &  & $55.0$ & $74.1$ & $58.3$ & $76.4$ \\
    \rowcolor{lightgray}{ViT-s +cascade} &  & $\mathbf{55.4}$ & $74.6$ & $\mathbf{58.9}$ & $\mathbf{76.7}$ \\
    \hline
    {ResNet-50 } & \multirow{6}*{object + part} &  $53.6$ & $74.1$ & $55.4$ & $76.4$ \\
    {ResNet-50 +cascade} &  & $54.2$ & $75.6$ & $60.9$ & $78.6$ \\
    {ResNet-101} &  & $57.6$ & $75.7$ & $60.5$ & $78.0$ \\
    {ResNet-101 +cascade} &  & $57.8$ & $75.9$ & $60.8$ & $78.7$ \\
    {ViT-s} &  & $57.9$ & $\mathbf{76.1}$ & $60.5$ & $78.1$ \\
    \rowcolor{lightgray}{ViT-s +cascade} &  & $\mathbf{58.4}$ & $\mathbf{76.1}$ & $\mathbf{61.3}$ & $\mathbf{79.0}$ \\
    \bottomrule
  \end{tabular}   
  \label{tab:objseg}
\end{table}

\noindent\textbf{Results.} To demonstrate that part annotations can enhance model performance for the object segmentation task, we first trained the aforementioned models using only object-level masks and bounding boxes. Subsequently, we trained the models using both object-level and part-level masks and bounding boxes. In \cref{tab:objseg}, under the Train Set column, `object' denotes the training set at the object level, while `object+part' indicates the training set incorporating both object-level and part-level annotations. 

\cref{tab:objseg} indicates that integrating part-level annotations into the training process significantly enhances the performance of all models across various metrics. This improvement is evident in both the Mask and Box columns, which correspond to instance segmentation and bounding box detection, respectively. For example, comparing the ResNet-50 FPN model trained with only object-level annotations (`object') to the same model trained with both object-level and part-level annotations (`object+part'), we observe a substantial improvement in performance metrics. 

\begin{figure}[t]
  \centering
   \includegraphics[width=\linewidth]{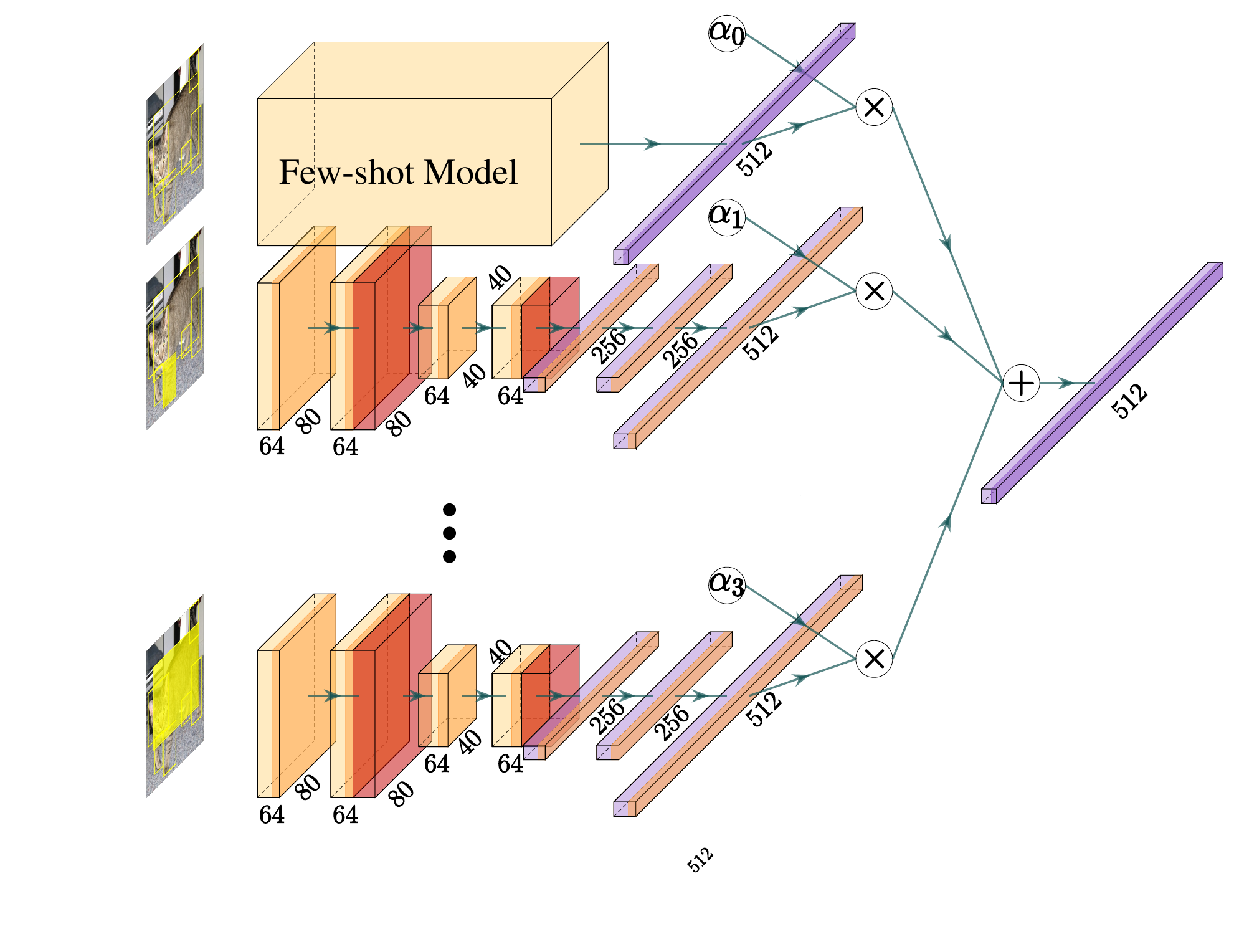}
   \caption{An overview of the framework for multi-branch feature extraction and normalized fusion. The topmost branch extracts features from the original image using a few-shot method (e.g., Meta-Baseline), while the subsequent three branches extract features from three distinct object parts. These four feature sets are then linearly combined, and their weights are normalized to produce the final representation.}
   \label{fig:fewshot}
\end{figure}

\section{PIN++ for Few-Shot Learning}
\label{sec:fewshot}

\noindent\textbf{Benchmark design.} For few-shot learning tasks, our focus was on determining the extent to which classification accuracy could be improved by models trained with part annotations, relative to those without such annotations. However, incorporating part information into previous models posed a significant challenge, as they were primarily trained without it. In our experiments, we employed two main methods: DeepEMD \cite{deepemd} and Meta-Baseline~\cite{metabaseline}. 

DeepEMD is a new few-shot image classification framework that leverages the Earth Mover's Distance (EMD) as a metric to compute structural distances between dense image representations for determining image relevance. In the updated version, DeepEMD V2, the authors found that instead of generating dense representations of images, it was better to randomly sample a set of regions in the images and only compute the EMD between these regions. Thus, it is natural to replace these randomly generated regions with annotated part bounding boxes.


We also evaluated Meta-Baseline as a few-shot learner. However, the original Meta-Baseline model, like most existing few-shot frameworks, could not be readily adapted to the part dataset through simple modifications—unlike DeepEMD. To address this limitation, we propose a novel framework that enables rapid adaptation of the original MetaBaseline architecture to the part dataset. 

Note that most few-shot classification models can be represented in the form of $h\circ g(x)$, where $x$ represents an image, $g$ is the backbone of the model used for extracting image features, and $h$ is the classification head of the model that maps these features to class labels. We can add $l$ part branches to the previous few-shot model, each with the same structure but independent parameters. Each part branch consists of several convolutional layers (Conv3x3) followed by ReLU functions and max pooling layers. These layers are designed to extract local features from the clipped regions of the image. For the input image $x$, clip out the bounding boxes of the $l$ largest parts by area, and then resize them to match the original image dimensions, denoted as $\{c_i\}$. If $x$ has no more than $l$ parts, we clip a few random rectangles from $x$. We input each $c_i$ into the $i$-th branch to extract the part features. The final image feature is composed of a linear combination of the $l$ part features and one image feature (obtained from $g$). The final model can be represented as: $$h(\sum\limits_{i=1}^{l}\alpha_i \cdot p_i(c_i) + (1-\sum\limits_{i=1}^{l}\alpha_i)\cdot g(x)),$$
where $p_i$ represents the $i$-th part CNN branch and $\alpha_i$ represents the weight of the $i$-th branch, which is a trainable parameter. In our experiments, we found a good trade-off between the number of model parameters and model performance for $l = 3$. 

\cref{fig:fewshot} shows the model architecture diagram for $l=3$ in \cref{fig:fewshot}. The image located at the top left is the original image. The three images positioned beneath it respectively contain the three largest parts of the cat, which are indicated by yellow masks. We then crop out these three parts. These are the inputs that the model will use to learn. The top box labeled ``Few-shot Method" indicates that the model is using some few-shot learning approach like Meta-Baseline, where the model is trained to adapt quickly to new tasks with limited data. The model has three branches, each processing a different part of the input image. Each branch consists of convolutional layers followed by max-pooling layers. These layers are used to extract features from the input images. After passing through the convolutional and max-pooling layers, the features are extracted and represented as 3D cubes. These features are then flattened into 1D vectors, which are ready for further processing. The vectors are then combined using trainable weights ($\alpha_0$, $\alpha_1$, $\alpha_2$, $\alpha_3$). These weights are applied element-wise to the vectors, indicating the importance of each feature in the final representation. Weight normalization is guaranteed by $\alpha_0=1-\alpha_1 - \alpha_2 - \alpha_3$. The final output is a single vector that represents the combined features from all three branches. This vector is used for the classification head.

Since our method does not modify the backbone, it is highly versatile and can be applied to various few-shot learning models, including those based on metric learning, meta-learning, and optimization-based approaches. This makes our proposed framework a valuable addition to the few-shot learning toolkit, providing a straightforward way to enhance model performance across different datasets. Another advantage of this design is that it enhances the interpretability of the model, allowing the contribution of each part to be analyzed according to the weight of each different branch.

\begin{table}[t]
\centering
\caption{Few-shot classification accuracy (\%) for the 5-way 1-shot task, comparing different methods and training data configurations (object only vs. object+part). }
\small
\begin{tabular}{c|c|c}
\toprule
{Method} & {Training Set} & {Accuracy} \\ 
\midrule
{DeepEMD} & object & $62.8$ \\ 
{DeepEMD} & object + part & ${64.0}$ \\
{Meta-Baseline} & object & $73.8$ \\
\rowcolor{lightgray}{Meta-Baseline} & object + part & $\mathbf{74.7}$ \\ 
 \bottomrule
\end{tabular}
\label{tab:fwos}
\end{table}

\begin{table}[t]
\centering
\caption{Few-shot classification accuracy (\%) for the 5-way 5-shot task, comparing different methods and training data configurations (object only vs. object+part). }
\small
\begin{tabular}{c|c|c}
\toprule
{Method} & {Training Set} & {Accuracy} \\
\midrule
{DeepEMD} & object & $77.8$ \\ 
 DeepEMD       & object + part & ${78.7}$ \\
{Meta-Baseline} & object & $90.6$ \\
\rowcolor{lightgray}Meta-Baseline    & object + part & $\mathbf{91.3}$ \\ 
 \bottomrule
\end{tabular}
\label{tab:fwfs}
\end{table}

\noindent\textbf{Results.} For the DeepEMD method, we employed part bounding boxes instead of random sampling. For the Meta-Baseline method, we utilized the model architecture illustrated in \cref{fig:fewshot}. The results in \cref{tab:fwos} demonstrate the 5-way 1-shot classification accuracy for different methods. Here the Meta-Baseline method outperforms DeepEMD, achieving higher classification accuracy when the training set includes only object-level information. \cref{tab:fwfs} shows the 5-way 5-shot classification accuracy. The results indicate a significant improvement in performance for both methods when part-level annotations are incorporated into the training process. Specifically, the Meta-Baseline method exhibits a substantial increase in classification accuracy, exceeding 90\% when trained with both object and part-level annotations. This underscores the advantages of incorporating part-level information in few-shot learning, as it equips the model with more detailed and nuanced features that are essential for accurate classification in data-scarce scenarios. 

\section{Conclusion}
\label{sec:conclusion}
In this study, we present PIN++, an extensive dataset that has been enhanced with detailed part annotations. This dataset is designed to drive research in part-based models for visual tasks. We introduce MPM, a pioneering part-supervised recognition model, specifically designed to fully leverage these annotations. This innovative approach has been demonstrated to enhance robustness across a multitude of scenarios and evaluation metrics, all without imposing additional computational costs during the inference phase. Our thorough benchmarking on PIN++ for a variety of visual tasks illuminates the dataset's versatile utility. The experimental results conclusively show the significant advantages of part annotations in bolstering performance for object segmentation, part segmentation, and few-shot learning tasks. We anticipate that the introduction of our PIN++ dataset will garner increased focus on the development of part-based models aimed at overcoming these challenges and revitalizing the significance of part-level analysis. 

\section{Declarations}

\subsection{Funding information}
This work was supported by the National Key Research and Development Program of China under Grant 2021ZD0200301, the National Natural Science Foundation of China under Grants U2341228 and 62576187, and the Fundamental and Interdisciplinary Disciplines Breakthrough Plan of the Ministry of Education of China.

\subsection{Competing interests}
The authors have no relevant financial or non-financial interests to disclose.

\subsection{Data availability}
\begin{itemize}
    \item \textbf{PartImageNet++:} \url{https://huggingface.co/datasets/lixiao20/PartImageNetPP}.
    \item \textbf{Other publicly available datasets used in this study: } 
    \begin{itemize}
        \item ImageNet \cite{imagenet}: \url{http://www.image-net.org/}
        \item Tiny-ImageNet \cite{tinyimagenet}: \url{https://github.com/jcjohnson/tiny-imagenet}
        \item PartImageNet \cite{partimagenet}: \url{https://huggingface.co/datasets/turkeyju/PartImageNet}
        \item PACO \cite{paco}: \url{https://github.com/facebookresearch/paco}
        \item Pascal-Part \cite{PASCAL-Part}: \url{https://zenodo.org/records/5878773#.YegfKiwo-qB}
        \item ADE20K \cite{ade20k}: \url{https://groups.csail.mit.edu/vision/datasets/ADE20K/}
        \item Cityscapes Panoptic-Parts \cite{Cityscapes-Panoptic-Parts}: \url{https://www.cityscapes-dataset.com}
        \item ImageNet-A-Plus \cite{in-a-plus}: \url{https://github.com/thu-ml/imagenet-a-plus}
        \item ImageNet-Sketch \cite{in-sketch}: \url{https://github.com/HaohanWang/ImageNet-Sketch}
        \item Stylized ImageNet \cite{sin}: \url{https://github.com/rgeirhos/Stylized-ImageNet}
        \item ImageNet Distortion Dataset \cite{modelvshuman}: \url{https://github.com/bethgelab/model-vs-human/}
    \end{itemize}
\end{itemize}

\bibliography{sn-bibliography}

@String(PAMI = {IEEE Trans. Pattern Anal. Mach. Intell. (TPAMI)})

@String(IJCV = {Int. J. Comput. Vis. (IJCV)})

@String(CVPR= {IEEE Conf. Comput. Vis. Pattern Recog. (CVPR)})

@String(ICCV= {Int. Conf. Comput. Vis. (ICCV)})

@String(ECCV= {Eur. Conf. Comput. Vis. (ECCV)})

@String(NIPS= {Adv. Neural Inform. Process. Syst. (NeurIPS)})

@String(BMVC= {Brit. Mach. Vis. Conf. (BMVC)})

@String(ACMMM= {ACM Int. Conf. Multimedia (ACM MM)})

@String(ICLR = {Int. Conf. Learn. Rep. (ICLR)})

@String(ICML = {Int. Conf. Mach. Learn. (ICML)})

@inproceedings{wideresnet,
  author       = {Sergey Zagoruyko and
                  Nikos Komodakis},
  title        = {Wide Residual Networks},
  booktitle    = BMVC,
  year         = {2016}
}

@inproceedings{generated,
  author       = {Sven Gowal and
                  Sylvestre{-}Alvise Rebuffi and
                  Olivia Wiles and
                  others},
  title        = {Improving Robustness using Generated Data},
  booktitle    = NIPS,
  pages        = {4218--4233},
  year         = {2021}
}

@inproceedings{stablediffusion,
  author       = {Robin Rombach and
                  Andreas Blattmann and
                  Dominik Lorenz and
                  Patrick Esser and
                  Bj{\"{o}}rn Ommer},
  title        = {High-Resolution Image Synthesis with Latent Diffusion Models},
  booktitle    = CVPR,
  pages        = {10674--10685},
  year         = {2022}
}

@article{diffusion,
  author       = {Qihao Liu and
                  Adam Kortylewski and
                  Yutong Bai and
                  Song Bai and
                  Alan L. Yuille},
  title        = {Intriguing Properties of Text-guided Diffusion Models},
  journal      = {arXiv preprint arXiv:2306.00974},
  year         = {2023}
}

@inproceedings{robustbench,
  author       = {Francesco Croce and
                  Maksym Andriushchenko and
                  Vikash Sehwag and
                  Edoardo Debenedetti and
                  Nicolas Flammarion and
                  Mung Chiang and
                  Prateek Mittal and
                  Matthias Hein},
  title        = {RobustBench: a standardized adversarial robustness benchmark},
  booktitle    = {NeurIPS Datasets and Benchmarks},
  year         = {2021}
}

@article{relu,
  author       = {Abien Fred Agarap},
  title        = {Deep Learning using Rectified Linear Units (ReLU)},
  journal      = {arXiv preprint arXiv:1803.08375},
  year         = {2018}
}

@article{gelu,
  title={Gaussian error linear units (gelus)},
  author={Hendrycks, Dan and Gimpel, Kevin},
  journal={arXiv preprint arXiv:1606.08415},
  year={2016}
}

@inproceedings{instanceasquery,
  author       = {Yuxin Fang and
                  Shusheng Yang and
                  Xinggang Wang and
                  Yu Li and
                  Chen Fang and
                  Ying Shan and
                  Bin Feng and
                  Wenyu Liu},
  title        = {Instances as Queries},
  booktitle    = ICCV,
  pages        = {6890--6899},
  year         = {2021}
}

@inproceedings{his1,
  author       = {Michael C. Burl and
                  Markus Weber and
                  Pietro Perona},
  title        = {A Probabilistic Approach to Object Recognition Using Local Photometry
                  and Global Geometry},
  booktitle    = ECCV,
  volume       = {1407},
  pages        = {628--641},
  year         = {1998}
}

@inproceedings{his2,
  author       = {Ross B. Girshick and
                  Forrest N. Iandola and
                  Trevor Darrell and
                  Jitendra Malik},
  title        = {Deformable part models are convolutional neural networks},
  booktitle    = CVPR,
  pages        = {437--446},
  year         = {2015}
}

@article{parthis,
  author       = {Martin A. Fischler and
                  Robert A. Elschlager},
  title        = {The Representation and Matching of Pictorial Structures},
  journal      = {{IEEE} Trans. Comput.},
  volume       = {22},
  number       = {1},
  pages        = {67--92},
  year         = {1973}
}

@article{advseg,
  author       = {Francesco Croce and
                  Naman D. Singh and
                  Matthias Hein},
  title        = {Robust Semantic Segmentation: Strong Adversarial Attacks and Fast
                  Training of Robust Models},
  journal      = NIPS,
  year         = {2023}
}

@inproceedings{zeroshot,
  title={Language-driven anchors for zero-shot adversarial robustness},
  author={Li, Xiao and Zhang, Wei and Liu, Yining and Hu, Zhanhao and Zhang, Bo and Hu, Xiaolin},
  booktitle=CVPR,
  pages={24686--24695},
  year={2024}
}

@article{advod,
  title={On the importance of backbone to the adversarial robustness of object detectors},
  author={Li, Xiao and Chen, Hang and Hu, Xiaolin},
  journal={IEEE Transactions on Information Forensics and Security},
  year={2025},
  publisher={IEEE}
}

@inproceedings{awp,
  author       = {Dongxian Wu and
                  Shu{-}Tao Xia and
                  Yisen Wang},
  title        = {Adversarial Weight Perturbation Helps Robust Generalization},
  booktitle    = NIPS,
  year         = {2020}
}

@inproceedings{trades,
  author    = {Hongyang Zhang and
               Yaodong Yu and
               Jiantao Jiao and
               Eric P. Xing and
               Laurent El Ghaoui and
               Michael I. Jordan},
  title     = {Theoretically Principled Trade-off between Robustness and Accuracy},
  booktitle = ICML,
  volume    = {97},
  pages     = {7472--7482},
  year      = {2019}
}

@inproceedings{AthalyeC018,
  author    = {Anish Athalye and
               Nicholas Carlini and
               David A. Wagner},
  title     = {Obfuscated Gradients Give a False Sense of Security: Circumventing Defenses to Adversarial Examples},
  booktitle = ICML,
  volume    = {80},
  pages     = {274--283},
  year      = {2018}
}

@inproceedings{adaptive20,
  author    = {Florian Tram{\`{e}}r and
               Nicholas Carlini and
               Wieland Brendel and
               Aleksander Madry},
  title     = {On Adaptive Attacks to Adversarial Example Defenses},
  booktitle = NIPS,
  year      = {2020}
}

@inproceedings{fasterrcnn,
  author       = {Shaoqing Ren and
                  Kaiming He and
                  Ross B. Girshick and
                  Jian Sun},
  title        = {Faster {R-CNN:} Towards Real-Time Object Detection with Region Proposal
                  Networks},
  booktitle    = NIPS,
  pages        = {91--99},
  year         = {2015}
}

@inproceedings{imagenetc,
  author    = {Dan Hendrycks and
               Thomas G. Dietterich},
  title     = {Benchmarking Neural Network Robustness to Common Corruptions and Perturbations},
  booktitle = ICLR,
  year      = {2019}
}

@inproceedings{glip,
  author       = {Liunian Harold Li and
                  Pengchuan Zhang and
                  Haotian Zhang and
                  Jianwei Yang and
                  Chunyuan Li and
                  Yiwu Zhong and
                  Lijuan Wang and
                  Lu Yuan and
                  Lei Zhang and
                  Jenq{-}Neng Hwang and
                  Kai{-}Wei Chang and
                  Jianfeng Gao},
  title        = {Grounded Language-Image Pre-training},
  booktitle    = CVPR,
  pages        = {10955--10965},
  year         = {2022}
}

@inproceedings{coco,
  title={Microsoft coco: Common objects in context},
  author={Lin, Tsung-Yi and Maire, Michael and Belongie, Serge and Hays, James and Perona, Pietro and Ramanan, Deva and Doll{\'a}r, Piotr and Zitnick, C Lawrence},
  booktitle=ECCV,
  pages={740--755},
  year={2014}
}

@inproceedings{opendet,
  author       = {Chuang Lin and
                  Peize Sun and
                  Yi Jiang and
                  Ping Luo and
                  Lizhen Qu and
                  Gholamreza Haffari and
                  Zehuan Yuan and
                  Jianfei Cai},
  title        = {Learning Object-Language Alignments for Open-Vocabulary Object Detection},
  booktitle    = ICLR,
  year         = {2023}
}

@inproceedings{at,
  title={Towards deep learning models resistant to adversarial attacks},
  author={Madry, Aleksander and Makelov, Aleksandar and Schmidt, Ludwig and Tsipras, Dimitris and Vladu, Adrian},
  booktitle    = ICLR,
  year={2018}
}

@inproceedings{naturaladv,
  author       = {Dan Hendrycks and
                  Kevin Zhao and
                  Steven Basart and
                  Jacob Steinhardt and
                  Dawn Song},
  title        = {Natural Adversarial Examples},
  booktitle    = CVPR,
  pages        = {15262--15271},
  year         = {2021}
}

@inproceedings{adv13,
  author       = {Christian Szegedy and
                  Wojciech Zaremba and
                  Ilya Sutskever and
                  Joan Bruna and
                  Dumitru Erhan and
                  Ian J. Goodfellow and
                  Rob Fergus},
  title        = {Intriguing properties of neural networks},
  booktitle    = ICLR,
  year         = {2014}
}

@article{infants,
title = {Object recognition and attention to object components by preschool children and 4-month-old infants},
author = {Robert A Haaf and Anne L Fulkerson and Brandon J Jablonski and Julie M Hupp and Stacey S Shull and Lisa Pescara-Kovach},
journal = {Journal of Experimental Child Psychology},
volume = {86},
number = {2},
pages = {108-123},
year = {2003},

}

@article{oldrbc,
  title={Family resemblances: Studies in the internal structure of categories},
  author={Rosch, Eleanor and Mervis, Carolyn B},
  journal={Cognitive Psychology},
  volume={7},
  number={4},
  pages={573--605},
  year={1975},
}

@article{attentioneye,
  title={The role of visual attention in saccadic eye movements},
  author={Hoffman, James E and Subramaniam, Baskaran},
  journal={Perception \& Psychophysics},
  volume={57},
  number={6},
  pages={787--795},
  year={1995},
}

@article{opc,
  title={Objects, parts, and categories.},
  author={Tversky, Barbara and Hemenway, Kathleen},
  journal={Journal of Experimental Psychology: General},
  volume={113},
  number={2},
  pages={169},
  year={1984},
}

@article{RBC,
  title={Recognition-by-components: a theory of human image understanding.},
  author={Biederman, Irving},
  journal={Psychological Review},
  volume={94},
  number={2},
  pages={115},
  year={1987},
}

@inproceedings{resnext,
  author       = {Saining Xie and
                  Ross B. Girshick and
                  Piotr Doll{\'{a}}r and
                  Zhuowen Tu and
                  Kaiming He},
  title        = {Aggregated Residual Transformations for Deep Neural Networks},
  booktitle    = CVPR,
  pages        = {5987--5995},
  year         = {2017}
}

@inproceedings{resnet,
  author       = {Kaiming He and
                  Xiangyu Zhang and
                  Shaoqing Ren and
                  Jian Sun},
  title        = {Deep Residual Learning for Image Recognition},
  booktitle    = CVPR,
  pages        = {770--778},
  year         = {2016}
}

@inproceedings{swin,
  author       = {Ze Liu and
                  Yutong Lin and
                  Yue Cao and
                  Han Hu and
                  Yixuan Wei and
                  Zheng Zhang and
                  Stephen Lin and
                  Baining Guo},
  title        = {Swin Transformer: Hierarchical Vision Transformer using Shifted Windows},
  booktitle    = ICCV,
  pages        = {9992--10002},
  year         = {2021}
}

@inproceedings{convnext,
  author       = {Zhuang Liu and
                  Hanzi Mao and
                  Chao{-}Yuan Wu and
                  Christoph Feichtenhofer and
                  Trevor Darrell and
                  Saining Xie},
  title        = {A ConvNet for the 2020s},
  booktitle    = CVPR,
  pages        = {11966--11976},
  year         = {2022}
}

@inproceedings{bairobust,
  author       = {Yutong Bai and
                  Jieru Mei and
                  Alan L. Yuille and
                  Cihang Xie},
  title        = {Are Transformers more robust than CNNs?},
  booktitle    = NIPS,
  pages        = {26831--26843},
  year         = {2021}
}

@inproceedings{transfer,
  author       = {Hadi Salman and
                  Andrew Ilyas and
                  Logan Engstrom and
                  Ashish Kapoor and
                  Aleksander Madry},
  title        = {Do Adversarially Robust ImageNet Models Transfer Better?},
  booktitle    = NIPS,
  year         = {2020}
}

@inproceedings{objparsing,
  author       = {Yifan Zhao and
                  Jia Li and
                  Yu Zhang and
                  Yonghong Tian},
  title        = {Multi-Class Part Parsing With Joint Boundary-Semantic Awareness},
  booktitle    = ICCV,
  pages        = {9176--9185},
  year         = {2019}
}

@article{ade20k,
  author       = {Bolei Zhou and
                  Hang Zhao and
                  Xavier Puig and
                  Tete Xiao and
                  Sanja Fidler and
                  Adela Barriuso and
                  Antonio Torralba},
  title        = {Semantic Understanding of Scenes Through the {ADE20K} Dataset},
  journal      = IJCV,
  volume       = {127},
  number       = {3},
  pages        = {302--321},
  year         = {2019}
}

@article{liu,
  title={A comprehensive study on robustness of image classification models: Benchmarking and rethinking},
  author={Liu, Chang and Dong, Yinpeng and Xiang, Wenzhao and Yang, Xiao and Su, Hang and Zhu, Jun and Chen, Yuefeng and He, Yuan and Xue, Hui and Zheng, Shibao},
  journal=IJCV,
  volume={133},
  number={2},
  pages={567--589},
  year={2025},
  publisher={Springer}
}

@inproceedings{debenedetti,
  title={A light recipe to train robust vision transformers},
  author={Debenedetti, Edoardo and Sehwag, Vikash and Mittal, Prateek},
  booktitle={First IEEE Conference on Secure and Trustworthy Machine Learning (SaTML)},
  pages={225--253},
  year={2023}
}

@misc{easyrobust,
  title={Easyrobust: A comprehensive and easy-to-use toolkit for robust computer vision},
  author={Mao, Xiaofeng and Chen, Yuefeng and Li, Xiaodan and Qi, Gege and Duan, Ranjie and Zhang, Rong and Xue, Hui},
  year={2022},
  howpublished = {\url{https://github.com/alibaba/easyrobust}},
}

@article{rock,
  author       = {Xiao Li and
                  Ziqi Wang and
                  Bo Zhang and
                  Fuchun Sun and
                  Xiaolin Hu},
  title        = {Recognizing Object by Components With Human Prior Knowledge Enhances
                  Adversarial Robustness of Deep Neural Networks},
  journal      = PAMI,
  volume       = {45},
  number       = {7},
  pages        = {8861--8873},
  year         = {2023}
}

@inproceedings{carlinipart,
  author       = {Chawin Sitawarin and
                  Kornrapat Pongmala and
                  Yizheng Chen and
                  Nicholas Carlini and
                  David A. Wagner},
  title        = {Part-Based Models Improve Adversarial Robustness},
  booktitle    = ICLR,
  year         = {2023}
}

@article{wang2015unsupervised,
  title={Unsupervised learning of object semantic parts from internal states of cnns by population encoding},
  author={Jianyu Wang and Zhishuai Zhang and Cihang Xie and Vittal Premachandran and Alan Yuille},
  journal={arXiv preprint arXiv:1511.06855},
  year={2015}
}

@inproceedings{CarFusion,
  author       = {N. Dinesh Reddy and
                  Minh Vo and
                  Srinivasa G. Narasimhan},
  title        = {CarFusion: Combining Point Tracking and Part Detection for Dynamic
                  3D Reconstruction of Vehicles},
  booktitle    = CVPR,
  pages        = {1906--1915},
  year         = {2018}
}

@inproceedings{Apollocar3D,
  author       = {Xibin Song and
                  Peng Wang and
                  Dingfu Zhou and
                  Rui Zhu and
                  Chenye Guan and
                  Yuchao Dai and
                  Hao Su and
                  Hongdong Li and
                  Ruigang Yang},
  title        = {ApolloCar3D: {A} Large 3D Car Instance Understanding Benchmark for
                  Autonomous Driving},
  booktitle    = CVPR,
  pages        = {5452--5462},
  year         = {2019}
}

@article{ATR,
  author       = {Xiaodan Liang and
                  Si Liu and
                  Xiaohui Shen and
                  Jianchao Yang and
                  Luoqi Liu and
                  Jian Dong and
                  Liang Lin and
                  Shuicheng Yan},
  title        = {Deep Human Parsing with Active Template Regression},
  journal      = PAMI,
  volume       = {37},
  number       = {12},
  pages        = {2402--2414},
  year         = {2015}
}

@article{LIP,
  author       = {Xiaodan Liang and
                  Ke Gong and
                  Xiaohui Shen and
                  Liang Lin},
  title        = {Look into Person: Joint Body Parsing {\&} Pose Estimation Network
                  and a New Benchmark},
  journal      = PAMI,
  volume       = {41},
  number       = {4},
  pages        = {871--885},
  year         = {2019}
}

@inproceedings{MHP,
  author       = {Jian Zhao and
                  Jianshu Li and
                  Yu Cheng and
                  Terence Sim and
                  Shuicheng Yan and
                  Jiashi Feng},
  title        = {Understanding Humans in Crowded Scenes: Deep Nested Adversarial Learning
                  and {A} New Benchmark for Multi-Human Parsing},
  booktitle    = ACMMM,
  pages        = {792--800},
  year         = {2018}
}

@inproceedings{CIHP,
  author       = {Ke Gong and
                  Xiaodan Liang and
                  Yicheng Li and
                  Yimin Chen and
                  Ming Yang and
                  Liang Lin},
  title        = {Instance-Level Human Parsing via Part Grouping Network},
  booktitle    = ECCV,
  volume       = {11208},
  pages        = {805--822},
  year         = {2018}
}

@inproceedings{PASCAL-Part,
  author       = {Xianjie Chen and
                  Roozbeh Mottaghi and
                  Xiaobai Liu and
                  Sanja Fidler and
                  Raquel Urtasun and
                  Alan L. Yuille},
  title        = {Detect What You Can: Detecting and Representing Objects Using Holistic Models and Body Parts},
  booktitle    = CVPR,
  pages        = {1979--1986},
  year         = {2014}
}

@article{Cityscapes-Panoptic-Parts,
  title={Cityscapes-panoptic-parts and PASCAL-panoptic-parts datasets for scene understanding},
  author       = {Panagiotis Meletis and
                  Xiaoxiao Wen and
                  Chenyang Lu and
                  Daan de Geus and
                  Gijs Dubbelman},
  journal={arXiv preprint arXiv:2004.07944},
  year={2020}
}

@inproceedings{partimagenet,
  author       = {Ju He and
                  Shuo Yang and
                  Shaokang Yang and
                  Adam Kortylewski and
                  Xiaoding Yuan and
                  Jieneng Chen and
                  Shuai Liu and
                  Cheng Yang and
                  Qihang Yu and
                  Alan L. Yuille},
  title        = {PartImageNet: {A} Large, High-Quality Dataset of Parts},
  booktitle    = ECCV,
  volume       = {13668},
  pages        = {128--145},
  year         = {2022}
}

@inproceedings{imagenet,
  author       = {Jia Deng and
                  Wei Dong and
                  Richard Socher and
                  Li{-}Jia Li and
                  Kai Li and
                  Li Fei{-}Fei},
  title        = {ImageNet: {A} large-scale hierarchical image database},
  booktitle    = CVPR,
  pages        = {248--255},
  year         = {2009}
}

@inproceedings{paco,
  author       = {Vignesh Ramanathan and
                  Anmol Kalia and
                  Vladan Petrovic and
                  Yi Wen and
                  Baixue Zheng and
                  Baishan Guo and
                  Rui Wang and
                  Aaron Marquez and
                  Rama Kovvuri and
                  Abhishek Kadian and
                  Amir Mousavi and
                  Yiwen Song and
                  Abhimanyu Dubey and
                  Dhruv Mahajan},
  title        = {{PACO:} Parts and Attributes of Common Objects},
  booktitle    = CVPR,
  pages        = {7141--7151},
  year         = {2023}
}

@article{vlpart,
  author       = {Peize Sun and
                  Shoufa Chen and
                  Chenchen Zhu and
                  Fanyi Xiao and
                  Ping Luo and
                  Saining Xie and
                  Zhicheng Yan},
  title        = {Going Denser with Open-Vocabulary Part Segmentation},
  journal      = {arXiv preprint arXiv:2305.11173},
  year         = {2023}
}

@article{maskrcnn,
  author       = {Kaiming He and
                  Georgia Gkioxari and
                  Piotr Doll{\'{a}}r and
                  Ross B. Girshick},
  title        = {Mask {R-CNN}},
  journal    = PAMI,
  volume       = {42},
  number       = {2},
  pages        = {386--397},
  year         = {2020},
}

@inproceedings{fpn,
  author       = {Tsung{-}Yi Lin and
                  Piotr Doll{\'{a}}r and
                  Ross B. Girshick and
                  Kaiming He and
                  Bharath Hariharan and
                  Serge J. Belongie},
  title        = {Feature Pyramid Networks for Object Detection},
  booktitle    = CVPR,
  pages        = {936--944},
  year         = {2017}
}

@inproceedings{sam,
    author    = {Kirillov, Alexander and Mintun, Eric and Ravi, Nikhila and Mao, Hanzi and Rolland, Chloe and Gustafson, Laura and Xiao, Tete and Whitehead, Spencer and Berg, Alexander C. and Lo, Wan-Yen and Dollar, Piotr and Girshick, Ross},
    title     = {Segment Anything},
    booktitle = ICCV,
    year      = {2023},
    pages     = {4015-4026}
}

@article{focal,
  author       = {Tsung{-}Yi Lin and
                  Priya Goyal and
                  Ross B. Girshick and
                  Kaiming He and
                  Piotr Doll{\'{a}}r},
  title        = {Focal Loss for Dense Object Detection},
  journal      = PAMI,
  volume       = {42},
  number       = {2},
  pages        = {318--327},
  year         = {2020}
}

@inproceedings{autoattack,
  author       = {Francesco Croce and
                  Matthias Hein},
  title        = {Reliable evaluation of adversarial robustness with an ensemble of
                  diverse parameter-free attacks},
  booktitle    = ICML,
  series       = {Proceedings of Machine Learning Research},
  volume       = {119},
  pages        = {2206--2216},
  year         = {2020}
}

@inproceedings{partseg,
  author       = {Shilong Liu and
                  Lei Zhang and
                  Xiao Yang and
                  Hang Su and
                  Jun Zhu},
  title        = {Unsupervised Part Segmentation Through Disentangling Appearance and
                  Shape},
  booktitle    = CVPR,
  pages        = {8355--8364},
  year         = {2021}
}

@article{in-a-plus,
  author       = {Xiao Li and
                  Jianmin Li and
                  Ting Dai and
                  Jie Shi and
                  Jun Zhu and
                  Xiaolin Hu},
  title        = {Rethinking Natural Adversarial Examples for Classification Models},
  journal      = {arXiv preprint arXiv:2102.11731},
  year         = {2021}
}

@inproceedings{liadbm,
  title={ADBM: Adversarial Diffusion Bridge Model for Reliable Adversarial Purification},
  author={Li, Xiao and Sun, Wenxuan and Chen, Huanran and Li, Qiongxiu and He, Yingzhe and Shi, Jie and Hu, Xiaolin},
  booktitle=ICLR,
  year={2025}
}

@inproceedings{li2025pbcat,
  title={PBCAT: Patch-based composite adversarial training against physically realizable attacks on object detection},
  author={Li, Xiao and Zhu, Yiming and Huang, Yifan and Zhang, Wei and He, Yingzhe and Shi, Jie and Hu, Xiaolin},
  booktitle=ICCV,
  year={2025}
}

@inproceedings{sin,
  author       = {Robert Geirhos and
                  Patricia Rubisch and
                  Claudio Michaelis and
                  Matthias Bethge and
                  Felix A. Wichmann and
                  Wieland Brendel},
  title        = {ImageNet-trained CNNs are biased towards texture; increasing shape
                  bias improves accuracy and robustness},
  booktitle    = ICLR,
  year         = {2019}
}

@inproceedings{in-sketch,
  author       = {Haohan Wang and
                  Songwei Ge and
                  Zachary C. Lipton and
                  Eric P. Xing},
  title        = {Learning Robust Global Representations by Penalizing Local Predictive
                  Power},
  booktitle    = NIPS,
  pages        = {10506--10518},
  year         = {2019}
}

@inproceedings{modelvshuman,
  author       = {Robert Geirhos and
                  Kantharaju Narayanappa and
                  Benjamin Mitzkus and
                  Tizian Thieringer and
                  Matthias Bethge and
                  Felix A. Wichmann and
                  Wieland Brendel},
  title        = {Partial success in closing the gap between human and machine vision},
  booktitle    = NIPS,
  pages        = {23885--23899},
  year         = {2021}
}

@article{cifar10,
  title={Learning multiple layers of features from tiny images},
  author={Krizhevsky, Alex and Hinton, Geoffrey and others},
  year={2009},
  publisher={Toronto, ON, Canada}
}

@article{tinyimagenet,
  title={Tiny imagenet visual recognition challenge},
  author={Le, Ya and Yang, Xuan},
  journal={CS 231N},
  volume={7},
  number={7},
  pages={3},
  year={2015}
}

@inproceedings{pang,
  author       = {Zekai Wang and
                  Tianyu Pang and
                  Chao Du and
                  Min Lin and
                  Weiwei Liu and
                  Shuicheng Yan},
  editor       = {Andreas Krause and
                  Emma Brunskill and
                  Kyunghyun Cho and
                  Barbara Engelhardt and
                  Sivan Sabato and
                  Jonathan Scarlett},
  title        = {Better Diffusion Models Further Improve Adversarial Training},
  booktitle    = ICML,
  volume       = {202},
  pages        = {36246--36263},
  year         = {2023}
}

@inproceedings{ema,
  author       = {Pavel Izmailov and
                  Dmitrii Podoprikhin and
                  Timur Garipov and
                  Dmitry P. Vetrov and
                  Andrew Gordon Wilson},
  title        = {Averaging Weights Leads to Wider Optima and Better Generalization},
  booktitle    = {{UAI}},
  pages        = {876--885},
  year         = {2018}
}

@inproceedings{labelsmoothing,
  title={Rethinking the inception architecture for computer vision},
  author={Christian Szegedy and
           Vincent Vanhoucke and
           Sergey Ioffe and
           Jonathon Shlens and
           Zbigniew Wojna},
  booktitle=CVPR,
  pages={2818--2826},
  year={2016}
}

@article{Mobile-Seed,
  author       = {Youqi Liao and
                  Shuhao Kang and
                  Jianping Li and
                  Yang Liu and
                  Yun Liu and
                  Zhen Dong and
                  Bisheng Yang and
                  Xieyuanli Chen},
  title        = {Mobile-Seed: Joint Semantic Segmentation and Boundary Detection for
                  Mobile Robots},
  journal      = {{IEEE} Robotics Autom. Lett.},
  volume       = {9},
  number       = {4},
  pages        = {3902--3909},
  year         = {2024},
}

@article{UDS,
  author       = {Jun Liu and
                  Junyuan Dong and
                  Mingming Hu and
                  Xu Lu},
  title        = {{UDS-SLAM:} real-time robust visual {SLAM} based on semantic segmentation
                  in dynamic scenes},
  journal      = {Ind. Robot},
  volume       = {51},
  number       = {2},
  pages        = {206--218},
  year         = {2024},
}

@article{3D-SeqMOS,
  author       = {Yuan Zhuang and
                  Qipeng Li and
                  Yiwen Chen and
                  Jianzhu Huai and
                  Miao Li and
                  Tianbing Ma and
                  Yufei Tang and
                  Xinlian Liang},
  title        = {3D-SeqMOS: {A} Novel Sequential 3D Moving Object Segmentation in Autonomous
                  Driving},
  journal      = {{IEEE} Trans. Intell. Transp. Syst.},
  volume       = {25},
  number       = {8},
  pages        = {8782--8795},
  year         = {2024}
}

@phdthesis{autoobj1,
  author       = {Nuri Benbarka},
  title        = {Instance Segmentation and 3D Multi-Object Tracking for Autonomous
                  Driving},
  school       = {T{\"{u}}bingen University, Germany},
  year         = {2023}
}

@inproceedings{vitdet,
  title={Exploring plain vision transformer backbones for object detection},
  author={Li, Yanghao and Mao, Hanzi and Girshick, Ross and He, Kaiming},
  booktitle=ECCV,
  pages={280--296},
  year={2022},
  organization={Springer}
}

@inproceedings{cascade,
  author       = {Zhaowei Cai and
                  Nuno Vasconcelos},
  title        = {Cascade {R-CNN:} Delving Into High Quality Object Detection},
  booktitle    = CVPR,
  year         = {2018}
}

@inproceedings{lsj,
  author       = {Golnaz Ghiasi and
                  Yin Cui and
                  Aravind Srinivas and
                  Rui Qian and
                  Tsung{-}Yi Lin and
                  Ekin D. Cubuk and
                  Quoc V. Le and
                  Barret Zoph},
  title        = {Simple Copy-Paste Is a Strong Data Augmentation Method for Instance
                  Segmentation},
  booktitle    = CVPR,
  year         = {2021},
}

@inproceedings{fedloss,
  author       = {Xizhou Zhu and
                  Weijie Su and
                  Lewei Lu and
                  Bin Li and
                  Xiaogang Wang and
                  Jifeng Dai},
  title        = {Deformable {DETR:} Deformable Transformers for End-to-End Object Detection},
  booktitle    = ICLR,
  year         = {2021},
}

@article{deepemd,
  author       = {Chi Zhang and
                  Yujun Cai and
                  Guosheng Lin and
                  Chunhua Shen},
  title        = {DeepEMD: Differentiable Earth Mover's Distance for Few-Shot Learning},
  journal      = PAMI,
  volume       = {45},
  number       = {5},
  pages        = {5632--5648},
  year         = {2023}
}

@inproceedings{metabaseline,
  author       = {Yinbo Chen and
                  Zhuang Liu and
                  Huijuan Xu and
                  Trevor Darrell and
                  Xiaolong Wang},
  title        = {Meta-Baseline: Exploring Simple Meta-Learning for Few-Shot Learning},
  booktitle    = ICCV,
  year         = {2021},

}

@article{xinlixue2,
  title={Human-level concept learning through probabilistic program induction},
  author={Brenden M. Lake and Ruslan Salakhutdinov and Joshua B. Tenenbaum},
  journal={Science},
  year={2015},
  volume={350},
  pages={1332 - 1338}
}

@inproceedings{pinpp,
  title={Partimagenet++ dataset: Scaling up part-based models for robust recognition},
  author={Li, Xiao and Liu, Yining and Dong, Na and Qin, Sitian and Hu, Xiaolin},
  booktitle=ECCV,
  pages={396--414},
  year={2024}
}

@article{fewshot1,
  author       = {Ju He and
                  Adam Kortylewski and
                  Alan L. Yuille},
  title        = {{COMPAS:} Representation Learning with Compositional Part Sharing
                  for Few-Shot Classification},
  journal      = {arXiv preprint arXiv:2101.11878},
  year         = {2021}
}

@inproceedings{fewshot2,
  author       = {Jiamin Wu and
                  Tianzhu Zhang and
                  Yongdong Zhang and
                  Feng Wu},
  title        = {Task-aware Part Mining Network for Few-Shot Learning},
  booktitle    = ICCV,
  year         = {2021},
}

@inproceedings{fewshot3,
  author       = {Weijian Xu and
                  Yifan Xu and
                  Huaijin Wang and
                  Zhuowen Tu},
  title        = {Attentional Constellation Nets for Few-Shot Learning},
  booktitle    = ICLR,
  year         = {2021},
}

@article{DMPCL,
  author       = {Mengya Han and
                  Yong Luo and
                  Han Hu and
                  Zengmao Wang and
                  Lefei Zhang and
                  Bo Du and
                  Ling{-}Yu Duan and
                  Dacheng Tao},
  title        = {{DM-PCL:} Text-Driven Dual-Modal Prototype Consistency Learning for
                  Weakly-Supervised Few-Shot Part Segmentation},
  journal      = IJCV,
  year         = {2025},
}

@article{PartWhole,
  author       = {Yi Liu and
                  Chengxin Li and
                  Shoukun Xu and
                  Jungong Han},
  title        = {Part-Whole Relational Fusion Towards Multi-Modal Scene Understanding},
  journal      = IJCV,
  year         = {2025},
}

@article{mambapw,
  author       = {Dingwen Zhang and
                  Liangbo Cheng and
                  Yi Liu and
                  Xinggang Wang and
                  Junwei Han},
  title        = {Mamba Capsule Routing Towards Part-Whole Relational Camouflaged Object
                  Detection},
  journal      = IJCV,
  volume       = {133},
  number       = {10},
  pages        = {7201--7221},
  year         = {2025}
}

@article{token,
  author       = {Zonghao Guo and
                  Fang Wan and
                  Mingxiang Liao and
                  Yidan Zhang and
                  Qixiang Ye},
  title        = {Discriminatively Matched Part Tokens for Pointly Supervised Instance
                  Segmentation},
  journal      = IJCV,
  year         = {2025},
}

\newpage
\onecolumn
\begin{appendices}
\renewcommand{\thefigure}{A\arabic{figure}}  
\renewcommand{\thetable}{A\arabic{table}}    
\setcounter{figure}{0}  
\setcounter{table}{0}
\section{Details on the Reused Annotations of PIN}
\label{sec:supp_reuse}

To reduce the annotation cost without compromising annotation quality, we selected and retained part annotations of some categories from the original PIN dataset \cite{partimagenet}, which originally consisted of 158 object categories. Our selection process involved several steps. First, we excluded all object categories in the PIN that belong to the super-categories \textit{snake}, \textit{car}, \textit{aeroplane}, and \textit{bottle}, since we found that the part annotations for these categories are insufficient for our purpose. For example, \textit{car} is annotated with \textit{side mirror}, \textit{body}, and \textit{tier} in PIN, whereas we required that \textit{window}, \textit{wheel}, \textit{front side}, \textit{left side}, \textit{right side}, \textit{back side}, and \textit{top side} should be annotated according to our annotation scheme. Second, for the remaining object categories belonging to the other super-categories, we carefully inspected their existing part annotations to see if they aligned with our specific annotation requirements. We further excluded categories like \textit{ram} where the existing annotations in PIN did not adequately capture important parts ($e.g.$, \textit{horn} that we considered significant). Lastly, we removed categories in PIN with less than 100 annotated images to ensure a sufficient amount of annotations for each category.

After applying the above procedures, we retained part annotations for a total of 90 object categories, which formed part of the PIN++ dataset. Below is a list of the retained object categories in PIN++:

\noindent n01440764, n01443537, n01484850, n01491361, n01494475, n01608432, n01614925, n01630670, n01632458, n01641577, n01644373, n01644900, n01664065, n01665541, n01667114, n01667778, n01669191, n01685808, n01687978, n01688243, n01689811, n01692333, n01693334, n01694178, n01695060, n01697457, n01698640, n01855672, n02002724, n02009229, n02009912, n02017213, n02025239, n02058221, n02071294, n02085782, n02089867, n02090379, n02092339, n02096177, n02096585, n02097474, n02098105, n02099601, n02100583, n02101006, n02101388, n02102040, n02102973, n02109525, n02109961, n02112137, n02114367, n02120079, n02124075, n02125311, n02128385, n02129604, n02130308, n02132136, n02133161, n02134084, n02134418, n02356798, n02397096, n02480495, n02480855, n02481823, n02483362, n02483708, n02484975, n02486261, n02486410, n02487347, n02488702, n02489166, n02490219, n02492035, n02492660, n02493509, n02493793, n02494079, n02514041, n02536864, n02607072, n02655020, n02835271, n03792782, n04482393, n04483307.

For the excluded 68 object categories, we reannotated them to ensure improved quality and alignment with our annotation scheme. Refer to \cref{fig:compare} for a visual comparison between our new annotations and the original annotations in PIN for some categories.

\section{Additional Part Segmentation Principles}
\label{sec:supp_principle}

In addition to the three principles described in \cref{sec:annotationscheme}, the following principles were followed to ensure high-quality part segmentation annotations. Firstly, the inclusion relation of parts (e.g., \textit{horn} is included in \textit{head}) should be annotated using a dictionary format. This dictionary specifies the relationship between the smaller internal part category (key) and the larger part category (value). It enables researchers to understand the hierarchical structure of the annotated part categories. Secondly, if there are multiple objects of the target category within an image, all of them should be annotated while all other objects are treated as background elements. Finally, some images may need to be discarded due to their low quality, but these discarded images should be retained to verify whether the reason for discarding meets our requirements, so as to avoid excluding difficult examples.

The annotation results, including annotated masks, inclusion relations, and records of discarded images, were returned to 317 in batches. If two out of ten randomly selected annotations for each category failed to meet the principles, the entire category’s annotations were rejected and re-annotated to ensure high-quality annotations. Annotations for one category may go through several iterations until they meet the requirements.

\section{Training Recipe Details}
\label{sec:supp_recipe}

\noindent\textbf{Standard training recipe}: The ResNet-50-GELU was optimized using stochastic gradient descent with an initial learning rate of 0.2, a momentum of 0.9, and a cosine decay scheduler for the learning rate. The weight decay was set to $1 \times 10^{-4}$. Data augmentation techniques, including random flipping and cropping, were applied during training. The model was trained for 80 epochs using 8 NVIDIA 3090 GPUs with a batch size of 512. The training process started with an initialization of a clean pretrained ResNet-50 checkpoint (on IN-1K) obtained from \texttt{torchvision}\footnote{\url{https://download.pytorch.org/models/ResNet-50-11ad3fa6.pth}}. Additionally, Exponential Moving Average (EMA) \cite{ema} with a decay of 0.9998 and label smoothing \cite{labelsmoothing} with a parameter of 0.1 were utilized. When training MPM, we used the same recipe for a fair comparison, except that we used an extra $L_{\rm{seg}}$ to introduce part-based supervision. Here $L_{\rm{seg}}$ used a Focal loss \cite{focal}, a variant of cross-entropy loss, to accelerate the convergence of part segmentation.

\noindent\textbf{Standard training recipe}: The standard training recipe closely resembles the AT recipe, except for the following differences. Instead of adversarial examples, clean examples are used during training. In addition, the models are trained for 100 epochs from scratch, without leveraging any pretrained checkpoints.

\section{Robustness on Corruptions and OOD Datasets}
\label{sec:supp_ood}
Apart from robustness against adversarial attacks that take into account the worst-case security threat scenarios, we evaluated the robustness of our part-based model on common image corruptions \cite{imagenetc} and several out-of-distribution (OOD) datasets such as \cite{in-a-plus, sin, in-sketch, modelvshuman}, which might pose more realistic threats. Given that adversarial training (AT) could have adverse effects on these non-adversarial scenarios as mentioned in \cite{rock, liu}, here we also evaluated models without AT. The training details are also presented in \cref{sec:setting}.

For common image corruptions \cite{imagenetc}, we generated various types of corrupted images on the entire IN-1K \texttt{val} set. The results of various models on these images are presented in \cref{tab:corruption}, where “Part” denotes MPM. The results of each corruption type are the accuracies averaged on five severity levels \cite{imagenetc}. It was observed that MPM exhibits superior robustness on almost all types of image corruptions compared to models lacking part supervision. In addition to common corruptions, we evaluated the models on four out-of-distribution datasets: ImageNet-A-Plus (IN-A-Plus) \cite{in-a-plus}, ImageNet Sketch (IN-Sketch) \cite{in-sketch}, Stylized ImageNet (SIN) \cite{sin}, and the image distortion dataset (DIN) \cite{modelvshuman}. IN-A-Plus consists of 3,286 images and serves as an improved version of ImageNet-A \cite{naturaladv}. IN-A-Plus comprises real-world and unmodified challenging images specifically curated for studying the robustness of classifiers to the internal variance of objects. IN-Sketch comprises 50,000 sketch-like images from the 1,000 categories in IN-1K. All images in this dataset are within a ``black and white'' color scheme. SIN is a stylized variant of IN-1K, where different styles of artistic paintings are randomly applied through style transfer techniques to the original images. We utilized the validation set of SIN. DIN comprises 18,080 OOD images, covering 17 different OOD distortion scenarios including changes to image texture and various forms of synthetic additive noise. The results shown in \cref{tab:odd-dataset} indicate that MPM also exhibits improved robustness on these OOD shifts.
\begin{table*}[!t]
  \centering
  \caption{Recognition accuracies (\%) of ResNet-50-Gelu on 15 different common image corruptions \cite{imagenetc}. We adopt accuracy as the metric to be consistent with other results, while this metric can be easily converted into the corruption error metric \cite{imagenetc}. ``Average'' denotes the accuracy averaged over different common corruptions. }
  \small
  \resizebox{\linewidth}{!}{
  \setlength{\tabcolsep}{0.7pt}
  {
    \begin{tabular}{cc|ccc|cccc|cccc|cccc|c}
    \multirow{2}*{Part} & \multirow{2}*{AT} & \multicolumn{3}{c|}{Noise}   & \multicolumn{4}{c|}{Blur} & \multicolumn{4}{c|}{Weather}  & \multicolumn{4}{c|}{Digital}& \multirow{2}*{Avg.}\\
                                     &                          & Gauss.  & Shot & \multicolumn{1}{c|}{Impul.}  & Defoc.  & Glass & Moti. & \multicolumn{1}{c|}{Zoom}  & Snow & Frost & Fog & \multicolumn{1}{c|}{Bright}  & Cont.  & Elast.  & Pixel  & JPEG \\
    \midrule
    $ $ & $ $  &$31.2$&$30.7$&$27.3$&$39.0$&$28.6$&$39.5$&$\mathbf{37.8}$&$35.8$&$40.7$&$53.8$&$69.0$&$38.0$&$\mathbf{37.0}$&$47.1$&$57.0$&$40.8$ \\
    $\checkmark$ & $ $ &$\mathbf{31.5}$&$\mathbf{31.0}$&$\mathbf{28.4}$&$\mathbf{41.9}$&$\mathbf{29.8}$&$\mathbf{40.3}$&$36.7$&$\mathbf{37.0}$&$\mathbf{41.3}$&$\mathbf{55.3}$&$\mathbf{69.6}$&$\mathbf{39.6}$&$36.9$&$\mathbf{52.9}$&$\mathbf{59.9}$&$\mathbf{42.1}$ \\
    \hline
    $ $ & $\checkmark$ &$29.6$&$28.5$&$21.0$&$23.3$&$32.9$&$32.8$&$34.4$&$34.2$&$34.0$&$10.8$&$58.4$&$9.5$&$52.2$&$58.8$&$63.1$&$34.9$ \\
    $\checkmark$ & $\checkmark$ &$\mathbf{30.1}$&$\mathbf{29.2}$&$\mathbf{22.4}$&$\mathbf{25.2}$&$\mathbf{35.1}$&$\mathbf{35.0}$&$\mathbf{36.4}$&$\mathbf{36.4}$&$\mathbf{36.2}$&$\mathbf{11.9}$&$\mathbf{59.7}$&$\mathbf{10.4}$&$\mathbf{53.1}$&$\mathbf{59.8}$&$\mathbf{64.3}$&$\mathbf{36.4}$ \\

    \end{tabular}
     }
     }
  \label{tab:corruption}%
\end{table*}%

\begin{table}[t]
  \centering
  \caption{Recognition accuracies (\%) of models on four OOD datasets.}
 \setlength{\tabcolsep}{4pt}
  {
    \begin{tabular}{cc|cccc|c}
    \toprule
     Part & AT  & IN-A-Plus & IN-Sketch & SIN & DIN  & Average  \\
     \midrule
     $ $ & $ $ & $6.9$ & $\mathbf{25.8}$ & $6.9$ & $53.5$ & $23.3$\\
     $\checkmark$ & $ $ & $\mathbf{7.4}$ & $25.7$ & $\mathbf{7.4}$ & $\mathbf{53.9}$ & $\mathbf{23.6}$\\
     \hline
     $ $ & $\checkmark$ & $5.3$ & $25.1$ & $12.5$ & $55.4$ & $24.6$\\
     $\checkmark$ & $\checkmark$ & $\mathbf{5.7}$ & $\mathbf{26.4}$ & $\mathbf{12.6}$ & $\mathbf{55.9}$ & $\mathbf{25.2}$\\
     \bottomrule
    \end{tabular}   
    }
  \label{tab:odd-dataset}
\end{table}

\section{Boosting Robustness on Downstream Tasks}
\label{sec:supp_downstream}

We conducted similar experiments by using the checkpoints of the vanilla baseline and MPM to initialize the backbone (a ResNet-50) of a Faster R-CNN and subsequently performing downstream AT using the recipe proposed by Li et al. \cite{advod}. The results on the downstream object detection are shown in \cref{tab:downstream}. MPM enhanced both the clean accuracy and adversarial robustness in object detection. Notably, there was a substantial boost in AP$_{50}$, a practical metric widely used in detection evaluations \cite{advod}. The results highlighted the importance of investigating part-based models on the large-scale IN-1K again.

\begin{table}[t]
  \centering
  \caption{Detection accuracies on MS-COCO \cite{coco}. $A_{\mathrm{cls}}$ and $A_{\mathrm{reg}}$ represent the attacks on the classification loss and regression loss of detectors, respectively. 
   }
  \small
 \setlength{\tabcolsep}{3pt}
  {
    \begin{tabular}{c|cc|cc|cc}
    \toprule
     \multirow{2}*{Initialization}  & \multicolumn{2}{c|}{Clean} & \multicolumn{2}{c|}{$A_{\mathrm{cls}}$}  & \multicolumn{2}{c}{$A_{\mathrm{reg}}$}  \\ 
             & AP  & \multicolumn{1}{c|}{AP$_{50}$}
             & AP  & \multicolumn{1}{c|}{AP$_{50}$}
             & AP  & \multicolumn{1}{c}{AP$_{50}$} \\
    \midrule
     \cite{advod} & $29.9$&$49.3$&$\mathbf{14.8}$&$25.5$&$19.7$&$40.5$\\
     \rowcolor{lightgray} ours (vanilla) &$29.9$&$49.8$&$14.5$&$25.4$&$19.3$&$40.2$\\
     \rowcolor{lightgray} ours (MPM) &$\mathbf{30.3}$&$\mathbf{50.3}$&$\mathbf{14.8}$&$\mathbf{25.9}$&$\mathbf{19.8}$&$\mathbf{40.7}$\\
     \bottomrule
    \end{tabular}   
    }
  \label{tab:downstream}
\end{table}

\section{Additional Ablation Studies}
\label{sec:supp_ablation}
\begin{table}[t]
  \centering
  \caption{Recognition accuracies (\%) of MPM on IN-1K trained with object segmentation mask and part segmentation mask.}
  \small

  \setlength{\tabcolsep}{4pt}
  {
    \begin{tabular}{c|cccc|c}
    \toprule
     Supervision & Clean & $l_\infty$ & $l_1$ & $l_2$ & Average  \\
     \midrule
     Object & $67.5$ & $37.3$ & $6.0$ & $23.6$ & $22.3$ \\
     Part & $\mathbf{67.8}$ & $\mathbf{39.1}$ & $\mathbf{6.2}$ &  $\mathbf{24.3}$ & $\mathbf{23.2}$ \\
     \bottomrule
    \end{tabular}
    }
  \label{tab:pl_ol}
\end{table}
We were interested in whether part segmentation labels could be replaced by simpler object segmentation labels. To explore this possibility, we initiated a study. Here we altered the model's supervision from part segmentation masks to their object-level counterparts, maintaining all other model elements constant. The results are presented in \cref{tab:pl_ol}. By comparing models trained with part segmentation masks against those trained with object segmentation masks, it was noted that the latter exhibited reduced accuracy rates for both unaltered and adversarially manipulated images. These outcomes underscore the critical value of employing detailed part-level annotations over general object segmentation annotations. 

\begin{table}[!t]
  \centering
  \caption{Recognition accuracies (\%) of different methods on T-IN. Previous methods used WideResNet-28-10 with the input resolution of $64 \times 64$ while we used ResNeXt-50 with the input resolution of $224 \times 224$. FLOPs and the number of parameters during inference are listed. $l_\infty$ and $l_\infty (\times 2)$ indicates $l_\infty$ attacks with the bounds $\epsilon_\infty = 4/255$ and $\epsilon_\infty = 8/255$, respectively. The original results with label leakage are shown in \textcolor{newgray}{gray}.}
  \small
 \setlength{\tabcolsep}{2pt}
  {
    \begin{tabular}{c|cc|ccc}
    \toprule
     Method & Param. & FLOPs & Clean &$l_\infty$ & $l_\infty (\times 2)$  \\
     \midrule
     Gowal et al. \cite{generated} & \multirow{3}*{$36.6$M} & \multirow{3}*{$21.0$G} & \textcolor{newgray}{$61.0$} & \textcolor{newgray}{-} & \textcolor{newgray}{$26.7$} \\
     Wang et al. \cite{pang} &  & & \textcolor{newgray}{$65.2$} & \textcolor{newgray}{$48.3$} & \textcolor{newgray}{$31.3$} \\
     Wang et al. \cite{pang} &  & & $57.6$ & $38.4$ & $28.4$\\
     \hline
     ours (vanilla) & \multirow{2}*{$23.4$M} & \multirow{2}*{$4.3$G} & $67.9$ & $48.0$ & $27.6$\\
     ours (MPM) &  & & $\mathbf{69.0}$ & $\mathbf{48.9}$ & $\mathbf{28.5}$ \\
     \bottomrule
    \end{tabular}   
    }
  \label{tab:tiny_compare}
\end{table}

\begin{figure}[!t]
  \centering
   \includegraphics[width=0.5\linewidth]{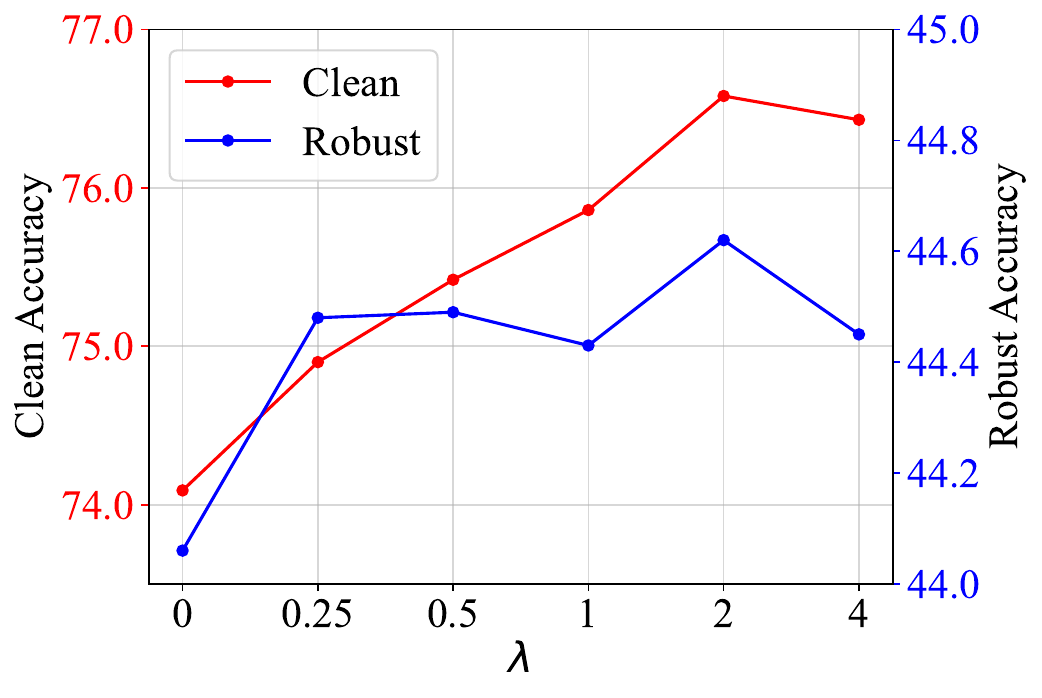}
   \caption{Recognition accuracies (\%) of MPM on T-IN with different $\lambda$. Here \textit{robust} is evaluated by the $l_\infty$ attack with a bound of $\epsilon = 4/255$.}
   \label{fig:tinyimagenet}
\end{figure}

It has come to our attention that the majority of research on adversarial robustness \cite{trades, awp, generated, pang} predominantly utilizes assessments on diminutive, low-resolution datasets such as CIFAR-10 \cite{cifar10}. To expand upon this, we execute experiments on an additional commonly referenced low-resolution dataset, Tiny-ImageNet (T-IN) \cite{tinyimagenet}. T-IN represents a subset of IN-1K, comprising 200 classes with 500 training samples per class, all adjusted to a resolution of 64×64 pixels. In parallel, we develop a corresponding dataset named Tiny-PartImageNet++ (T-PIN++), leveraging the publicly available generation code\footnote{\url{https://github.com/jcjohnson/tiny-imagenet}} of T-IN. T-PIN++ features part annotations (or pseudo part labels) for every training image and mirrors T-IN in terms of categories and training images, yet it refrains from resizing, thereby preserving the original high-resolution quality of the IN-1K images. 

Utilizing the AT recipe outlined in \cref{sec:setting}, we train both the standard ResNeXt-50 and MPM, with ResNeXt-50 \cite{resnext} serving as the foundational architecture, on the T-PIN++ dataset. Subsequently, we juxtapose these models against SOTA methodologies on T-IN \cite{generated, pang}, where WideResNet-28-10 \cite{wideresnet} is trained with an additional 1M images synthesized via diffusion models \cite{diffusion}. The comparative outcomes are documented in \cref{tab:tiny_compare}. It is noteworthy that the validation subset of T-IN originates from the training subset of IN-1K, resulting in label leakage during the assessment of prior SOTA techniques \cite{generated, pang}, given that the diffusion models employed are trained on the complete IN-1K training subset. 
The initial findings of these methods are indicated in gray within \cref{tab:tiny_compare}. We reassess the performance on the IN-1K \texttt{val} subset using the publicly available checkpoint from Wang et al. \cite{pang} (Gowal et al. \cite{generated} are not reassessed due to the unavailability of their released checkpoint). The analysis reveals that MPM, operating at an input resolution of 224×224, outperforms preceding SOTA approaches, which function at a 64×64 resolution and entail significantly more substantial models, on T-IN in terms of both adversarial robustness and clean accuracy. Moreover, MPM accomplishes these enhanced results with approximately one-fifth the computational FLOPs. These findings underscore the importance of examining adversarial robustness on high-resolution imagery, instead of being confined to low-resolution datasets such as CIFAR-10 or T-IN. 

Furthermore, we explore the impact of the hyper-parameter $\lambda$ (refer to the training objective detailed in \cref{sec:modeldesign}) on MPM's performance. A $\lambda$ value of 0 equates to the standard model devoid of part supervision, while a higher $\lambda$ signifies a greater focus on part supervision. Due to constraints in computational capacity, the ablation study is conducted using ResNet-50 on T-PIN++ (as previously described). The recognition accuracies of MPM for varying $\lambda$ values are depicted in \cref{fig:tinyimagenet}. It is evident that as $\lambda$ escalates, so does the clean accuracy of MPM. Moreover, the adversarial robustness of MPM exhibits minimal sensitivity to the precise $\lambda$ value, provided that $\lambda$ is greater than zero. Overall, these findings underscore the significance of incorporating part supervision. 

\end{appendices}



\end{document}